\documentclass[pdflatex,sn-mathphys-num]{sn-jnl}
\usepackage{graphicx}%
\usepackage{multirow}%
\usepackage{amsmath,amssymb,amsfonts}%
\usepackage{amsthm}%
\usepackage{mathrsfs}%
\usepackage[title]{appendix}%
\usepackage{xcolor}%
\usepackage{textcomp}%
\usepackage{manyfoot}%
\usepackage{booktabs}%
\usepackage{algorithm}%
\usepackage{algorithmicx}%
\usepackage{svg}
\usepackage{algpseudocode}%
\usepackage{listings}%
\usepackage{amsmath}      
\usepackage{float}
\usepackage{booktabs}
\usepackage{soul}
\usepackage{natbib}
\usepackage{adjustbox}
\usepackage{caption}
\usepackage{float}
\usepackage{booktabs}
\usepackage{bm}      
\theoremstyle{thmstyleone}%

\usepackage{listings}
\usepackage{anyfontsize}
\usepackage{xcolor}
\usepackage{lmodern}
\usepackage{caption}

\theoremstyle{thmstyletwo}%

\theoremstyle{thmstylethree}%

\begin{document}

\title{FuDoBa: Fusing Document and Knowledge Graph based Representations with Bayesian Optimisation}
\author*[1]{\fnm{Boshko} \sur{Koloski}}\email{boshko.koloski@ijs.si}

\author[1]{\fnm{Senja} \sur{Pollak}}\email{senja.pollak@ijs.si}

\author[3]{\fnm{Roberto} \sur{Navigli}}\email{navigli@diag.uniroma1.it}

\author*[1]{\fnm{Bla\v{z}} \sur{\v{S}krlj}}\email{blaz.skrlj@ijs.si}

\affil[1]{\orgname{Jo\v{z}ef Stefan Institute}, \orgaddress{\city{Ljubljana}, \country{Slovenia}}}

\affil[2]{\orgname{Jo\v{z}ef Stefan Postgraduate School}, \orgaddress{\city{Ljubljana}, \country{Slovenia}}}

\affil[3]{\orgname{Sapienza NLP Group, Sapienza University of Rome}, \orgaddress{\city{Rome}, \country{Italy}}}

\abstract{Building on the success of Large Language Models (LLMs), LLM-based representations have dominated the document representation landscape, achieving great performance on the document embedding benchmarks. However, the high-dimensional, computationally expensive embeddings from LLMs tend to be either too generic or inefficient for domain-specific applications. To address these limitations, we introduce FuDoBa—a Bayesian optimisation-based method that integrates LLM-based embeddings with domain-specific structured knowledge, sourced both locally and from external repositories like WikiData. This fusion produces low-dimensional, task-relevant representations while reducing training complexity and yielding interpretable early-fusion weights for enhanced classification performance. We demonstrate the effectiveness of our approach on six datasets in two domains, showing that when paired with robust AutoML-based classifiers, our proposed representation learning approach performs on par with, or surpasses, those produced solely by the proprietary LLM-based embedding baselines.}


\maketitle

\section{Introduction}\label{sec:intro}

Efficient and rich document representations are the building blocks for many natural language processing (NLP) tasks such as classification or clustering \cite{mteb}. Contemporary methods for representing documents focus on distilling representations from either pre-trained language models (PLMs) such as BERT \cite{devlin} or large language models (LLMs) such as Llama3 \cite{lamma3}, exploiting the rich semantic knowledge acquired during pre-training on vast text corpora. For instance, Sentence-BERT \cite{sbert} builds document representation by pooling over pre-trained BERT-based word embeddings, which are further refined through contrastive learning and Siamese networks. Similarly, LLM2Vec \cite{llm2vec} disentangles the causal masking of LLMs to a bi-directional one, further post-training the LLM on a masked next token prediction task and finally, training with a contrastive training objective, similarly to Sentence-BERT, refining the final representations via mean pooling by training with a contrastive training objective.

Despite good performance on public benchmarks such as MTEB ~\cite{mteb}, contrastive pre-training models require acquiring a dataset of triplet sentences (i.e., query, positive answer, and negative answer), which is often infeasible and costly. That said, even assuming such a dataset is available, training is costly and requires extra compute. The best-performing representations to date often come from proprietary companies and their mechanisms are largely unknown \cite{mteb}. Current approaches face two notable limitations: first, these embeddings typically reside in high-dimensional spaces, often exceeding 1,000 dimensions, creating practical challenges for efficient storage and computation (especially for AutoML model training); second, their generalist nature makes them suboptimal for domain-specific applications without expensive fine-tuning, which in some cases is infeasible.

One way to overcome the costly training requirements while still leveraging the expressive power of PLM/LLM derived representations is to introduce additional knowledge from auxiliary representations. Previous studies proposed representation enrichment via additional representations, either via representation alignment \cite{zero_shot_kg} or representation fusion \cite{KOLOSKI2022208}. In many cases, the approach of introducing auxiliary knowledge to document representations and searching over a space of possible classifiers with AutoML performs on par with task-specific fine-tuned PLMs such as BERT, without training costly classifiers \cite{autobot}.

Recently,~\citet{babelfusion} proposed BabelFusion, involving the introduction of \textit{global knowledge} from a knowledge graph by entity-linking to BabelNet's \cite{babelnet} synsets connected to a subgraph of WikiData5m \cite{wikidata}. They showed that projecting the inputs into low-dimensions not only produces low-dimensional spaces but can generate models that usually out-perform the language-only representations in the high dimensional space. However, they concluded that matching datasets where text stems from online discussion and is usually manifested with short texts with against a knowledge graph was sub-optimal. Indeed, this proved deteriorating for the downstream performance of AutoML classifiers, in comparison to that of weak classifiers fitted on the original space. As an alternative to their approach, we hypothesise that extracting domain-specific knowledge graphs, resulting in a \textit{local knowledge} graph\footnote{We refer to them as LocKG throughout the remainder of the paper.}, can provide local context that could complement to a global knowledge graph such as WikiDAta, aiding LLMs downstream.  

\begin{figure}[H]
    \centering
    \resizebox{0.9\textwidth}{!}{\includegraphics{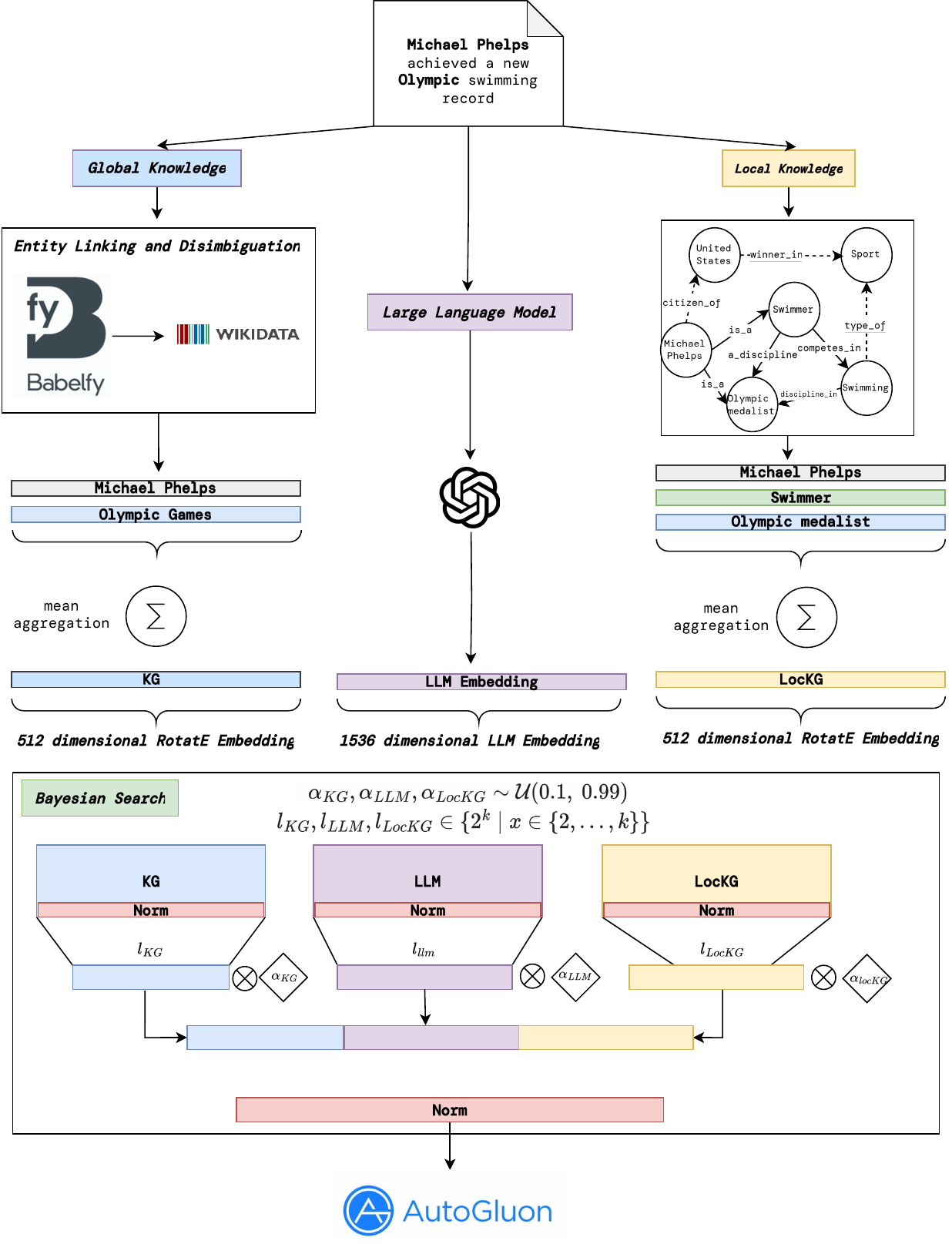}}
    \caption{Overview of our proposed \textit{FuDoBa} framework for representation enrichment. Starting from an LLM-based embedding (purple) external knowledge is incorporated via two parallel pathways: a global knowledge graph (light blue) branch that links entities using Babelfy~\cite{babelfy} and WikiData5m~\cite{wikidata} embeddings, and a novel local knowledge graph branch (yellow) that constructs domain-specific knowledge-graph via relation-extraction models. Representations are projected into a lower-dimensional space ($l$) and weighted by an importance parameter ($\alpha$), with ElasticNet normalisation applied both before and after fusion. The concatenated representation is then processed by AutoGluon~\cite{agtabular} AutoML for model search under a constrained time budget.}

    \label{fig:main-schema}
\end{figure}

In this work, we propose \textit{FuDoBa} (Figure~\ref{fig:main-schema}), a novel Bayesian optimisation-based early fusion methodology that builds on the idea of \citet{babelfusion}. It combines the semantic richness of LLM representations with the structured information from knowledge graphs and proposes the construction of local knowledge by extracting structured relations (e.g., knowledge triplets) from the dataset using Relation Extraction methods. Our approach systematically leverages external knowledge sources such as WikiData alongside these domain-specific, locally extracted knowledge structures to create more contextualised and task-relevant document representations. By employing Bayesian optimisation techniques, we identify importance parameters that maximize performance while minimising dimensionality. Furthermore, the optimised global weights assigned by Bayesian optimisation to the different representation sources serve as a a task-specific interpretation of the embeddings' importance. 

The main contributions of this work are as follows:
\begin{itemize}
    \item We demonstrate that LLM-based embeddings can be further improved by integrating both global contextual information and fine-grained, domain-specific local knowledge. In particular, we systematically evaluate the impact of constructing and incorporating a domain-specific knowledge graph—capturing domain dependant relations between entities.
    \item We experimentally demonstrate that training AutoML in low-dimensional spaces yields downstream models that perform as well as—or even better than—those produced using conventional methods.
    \item We introduce a novel, Bayesian optimisation–based method for low-dimensional early fusion of document and knowledge graph representations. 
\end{itemize}

The remainder of this paper is structured as follows. Section~\ref{sec:related_work} reviews the relevant literature, and Section~\ref{sec:methodology} outlines the proposed methodology. Section~\ref{sec:exp} describes the experimental setup, and Section~\ref{sec:results} presents the results. Section~\ref{sec:discussion} discusses the findings, and Section~\ref{sec:conc} concludes the paper. The Appendix details the method’s limitations in Section~\ref{abl:limitations}, implementation specifics in Section~\ref{abl:impl_details}, a deeper analysis of relation extractors within the local knowledge graph in Section~\ref{abl:rel_ext}, and additional experimental results in Section~\ref{apn:compl}.

\section{Related Work}\label{sec:related_work}
The task of document representation has evolved from pre-training shallow neural networks (NNs) on web scale data~ \cite{le-mikolov-2014-distributed} through transformer encoders \cite{devlin} refined by contrastive learning \cite{sbert, gao-etal-2021-simcse, li-li-2024-aoe}, to adaptations of decoder-only LLMs \cite{aghajanyan-etal-2024-llm2vec, khosla-etal-2025-magnet} and synthetic data approaches \cite{wang-etal-2024-improving-text}. While these techniques output powerful high-performant embeddings \cite{mteb}, significant challenges remain \cite{babelfusion}. Training and adapting these large models to novel domains requires substantial computational resources \cite{sbert}. Data acquisition remains laborious and potentially costly \cite{wang-etal-2024-improving-text}.
To avoid costly retraining, researchers have proposed utilising additional representations in LLM-based document representation. \citet{zero_shot_kg} enhanced document classification through zero-shot alignment between labels and knowledge graph concepts. Recent work~\cite{rita} demonstrated that adding reflection tokens to vision-language models with Wikipedia improves visual question answering performance. Tax2Vec \cite{tax2vec} leveraged the WordNet taxonomy to enhance TF-IDF features for short document classification. Koloski et al. \cite{KOLOSKI2022208} introduced Sentence-Transformers \cite{sbert}, sub-symbolic and WikiData5m \cite{wikidata} knowledge-graph embeddings, improving fake news classification over language-only representations. \citet{ostendorff2019enriching} proposed early-fusing a knowledge graph with BERT\cite{devlin} representations and refining them via a 2-layer MLP, showing remarkable results for multi-class classification. The benefits of local knowledge graphs (LocKG) have been noted in the context of information retrieval by~\citet{hybridrag}, who investigated how LocKG improves downstream performance in document search and retrieval. Similarly,~\citet{loc_glob} applied LocKG to the task of multi-document summarization. However, all of the mentioned approaches operate in high-dimensional spaces, presenting challenges due to the curse of dimensionality \cite{curse}.
Projecting these representations into low-dimensional space via SVD \cite{core} by preserving top-k singular values can compress dimensionality and facilitate representation fusion \cite{babelfusion, svd_fusion}. For efficient learning in these spaces, automated machine learning approaches like AutoGluon \cite{agtabular} excel on benchmarks through dual optimisation of parameters and model ensembles. While previous work \cite{babelfusion} combines low-dimensional fusion of knowledge and text utilised tree-based models TPOT~\cite{tpot}, our work focuses on incorporating AutoGluon \cite{agtabular}, a stronger model which also searches neural models. As this work connects to the field of data fusion, particularly early-fusion, we refer interested readers to~\cite{fusion1,fusion2} for further details. The goal of this study is to explore whether, by employing modality-specific low-dimensional projection---including local and global knowledge graphs and utilising strong AutoML models---we can perform on par, or similarly to original high-dimensional embeddings. In Table ~\ref{tab:method_comparison}, we compare our work with similar investigations. Notably, our approach introduces domain constructed local knowledge graphs, utilises stronger AutoML models, introduces modality specific intrepretable weights, and does not require access to embedding model weights nor requires custom models. 

\begin{table}[t]
\centering
\resizebox{\textwidth}{!}{
\begin{tabular}{lcccc}
\hline
\textbf{Feature} & \textbf{Text-KG}  & \textbf{Het. Ens.}  & \textbf{BabelFusion} & \textbf{FuDoBa} \\
& \textbf{Alignment} \cite{zero_shot_kg} &\cite{KOLOSKI2022208} & \cite{babelfusion}  & \textbf{(This Work)} \\
\hline
Model Weights & $\checkmark$ & $\times$ & $\times$ & $\times$ \\
\hline
Word-Disambiguation & $\times$ & $\times$ & BabelFy~\cite{babelnet}& BabelFy~\cite{babelnet}\\
\hline
Low-Dimensional Space & $\times$ & $\times$  & $\checkmark$ & $\checkmark$  \\
\hline
Modality importance & $\times$ & $\times$ & $\times$ & $\checkmark$ \\
\hline
Architecture & Zero-shot(*) & Custom NN & TPOT~\cite{tpot} & AutoGluon~\cite{agtabular} \\ \hline
External KG& ConceptNet~\cite{conceptnet} & WikiData5m~\cite{wikidata} & WikiData5m~\cite{wikidata}  & WikiData5m~\cite{wikidata}  \\ \hline
Local KG & $\times$ & $\times$ & $\times$ & $\checkmark$ \\ \hline
\hline
\end{tabular}
}
\caption{Comparison of different methods for leveraging knowledge graphs for LLM representation improvement.}
\label{tab:method_comparison}
\end{table}

\section{Methodology}\label{sec:methodology}
We introduce \textbf{FuDoBa}, a novel multimodal fusion framework that employs Bayesian optimisation to determine the optimal contribution of each modality’s low-dimensional projection for improved classification performance. 

\subsection{Document Representations}\label{subs:rep}
Our approach builds on the BabelFusion representations~\cite{babelfusion} by using LLM- and KG-based representations as the backbone for document representation.~\citet{babelfusion} observed that mapping textual content to a general-domain KG sometimes results in only a few matched entities for specific domains.  To address this limitation, the solution our approach proposes is the construction of a local KG (LocKG) using relation extraction techniques. Our framework integrates three complementary embedding modalities:

\begin{itemize} 
\item \textbf{Text Embeddings:} Embeddings are derived from OpenAI’s \texttt{text-embedding-2-small} model and capture rich contextual information from text. They are represented in 1536 dimensions. 
\item \textbf{Knowledge Graph (KG) Embeddings:} Generated via BabelFusion~\cite{babelfusion}, these embeddings map textual terms to Wikidata5M~\cite{wikidata} entities using Babelfy~\cite{babelfy}'s entity linking and word sense disambiguation and aggregate them with 512-dimensional RotatE~\cite{sun2018rotate} embeddings. 
\item \textbf{Local Knowledge Graph (LocKG) Embeddings:} These embeddings are obtained by extracting knowledge triplets using relation extractors and then embedding them with 512-dimensional RotatE~\cite{sun2018rotate} embeddings to capture localised relational structures. In our experiments, we employ prompt-guided \texttt{gpt-4o-mini} as an extractor.\footnote{Implementation details are present in Appendix~\ref{abl:impl_details}. In the Appendix ~\ref{abl:rel_ext} we show that this extractor can be changed for a smaller, open-sourced one, achieving similar results.}
\end{itemize}

\subsection{Preliminaries}

Two key techniques further underpin our pipeline:
\begin{itemize}
    \item \textbf{Dimensionality Reduction via Truncated SVD:} Given a data matrix $\bm{A} \in \mathbb{R}^{N \times d}$, its truncated SVD~\cite{svd} retains the top $p$ singular components: $\bm{A} \approx \bm{U}_p \bm{\Sigma}_p \bm{V}_p^T$, where $\bm{U}_p \in \mathbb{R}^{N \times p}$, $\bm{\Sigma}_p \in \mathbb{R}^{p \times p}$ (diagonal), and $\bm{V}_p \in \mathbb{R}^{d \times p}$. The columns of $\bm{V}_p$ represent the principal directions of variation in the column space. Projecting the original data onto these directions yields the reduced-dimension representation $\bm{P} = \bm{A} \bm{V}_p \in \mathbb{R}^{N \times p}$, which captures the maximal variance in $p$ dimensions. 
    
    \item \textbf{Normalisation:} We employ normalisation techniques before and after fusion. The $L_k$-norm of a vector $\bm{x} \in \mathbb{R}^d$ is defined as $\|\bm{x}\|_k = (\sum_{i=1}^d |x_i|^k)^{1/k}$. We utilize Elastic Net style normalisation, defined for a vector $\bm{x}$ as:
    \begin{equation*}
    \label{eq:elastic_net_norm_generic}
    N(\bm{x}) = \frac{\bm{x}}{w_1 \|\bm{x}\|_1 + w_2 \|\bm{x}\|_2},
    \end{equation*}
    where $w_1, w_2 \ge 0$ are weighting factors (e.g., $w_1=w_2=0.5$). This balances the regularising effects of L1 and L2 norms.
\end{itemize}

\subsection{FuDoBa: Fusion Mechanism and Parameter Optimisation}
\label{sec:FuDoBa_mechanism_opt}

\citet{babelfusion} demonstrated that projecting KGs  enriched high-dimensional LLM-embeddings to lower-dimensions have a comparable or better performance than base representations. However, their approach involved concatenating modalities before projection and weighting features with equal weights post-projection. We argue that by doing so information was weighted equally, which can impact performance while projecting, as different signals might be important for different tasks. We propose a different strategy where each modality's embedding, $\bm{E}_m$ ($m \in \{\text{llm}, \text{kg}, \text{loc}\}$), is first projected independently via Truncated SVD~\cite{svd} to a lower dimension $l_m$, yielding $\bm{P}_m = \operatorname{SVD}_{p_m}(\text{Normalize}(\bm{E}_m))$. This step aims to capture the most important information from each modality separately. Subsequently, each projection $\bm{P}_m$ is scaled by a factor $\alpha_m$ before concatenation. We hypothesise that the scaling factor $\alpha_m$ per modality can act as a learned global importance weight for the modality's contribution within the fusion, potentially improving performance and offering insights into the relative importance of modalities for a given task. \footnote{N.B. when a $\alpha$ is equal to \textbf{zero (0.0)}, the modality is excluded from the learning process as uninformative.} 

The fusion process is governed by a hyper-parameter vector $\bm{\theta}$, which comprises the projection dimensions and scaling factors for all modalities. For example, in the case of the LLM modality, $l_{\text{llm}}$ denotes the projection dimension, while $\alpha_{\text{llm}}$ represents the corresponding importance weight.: 
\begin{equation*}
\label{eq:theta_fusion_def}
\bm{\theta} =
\underbrace{\left(l_{\text{llm}},\, l_{\text{kg}},\, l_{\text{loc}}\right)}_{\text{projection dimensions}}
,\underbrace{\left(\alpha_{\text{llm}},\, \alpha_{\text{kg}},\, \alpha_{\text{loc}}\right)}_{\text{modality importance}}.
\end{equation*}
We define the search space $\Theta$ for these hyper-parameters as follows: each projection dimension $l_m$ is chosen from the set $\{16,32,64\}$, and each scaling multiplier $\alpha_m$ is selected from the discrete set $\{0, 0.1, \dots, 0.9, 1\}$.

\begin{algorithm}[ht]
\caption{FuDoBa: Low-Dimensional Multi-modal Embedding Fusion}
\label{alg:FuDoBa}
\begin{algorithmic}[1]
\Require Modality Embeddings $\{\bm{E}_m\}_{m\in\mathcal{M}}$ ($\mathcal{M}=\{\text{llm},\text{kg},\text{loc}\}$), Labels $\bm{y}$ 
\Require Importance weights $\{\alpha_m\}_{m\in\mathcal{M}}$ and Projection dimensions $\{l_m\}_{m\in\mathcal{M}}$
\Require AutoGluon classifier $\mathcal{C}$, Cross-Validation folds $k$ 
\Ensure Mean CV Macro F1-score $f$, Final classifier $\mathcal{C}$ (fit on full data)

\Statex \textit{ --- Step 1: Generate Fused Representation --- }
\For{$m \in \mathcal{M}$}
    \State $\bm{E}_m^{\text{norm}} \gets N(\bm{E}_m)$ \Comment{Elastic Net normalisation}
    \State $\bm{P}_m \gets \operatorname{SVD}_{p_m}(\bm{E}_m^{\text{norm}})$ \Comment{Truncated SVD to dimension $l_m$}
    \State $\bm{S}_m \gets \alpha_m \cdot \bm{P}_m$ \Comment{Scale projection by importance weight $\alpha_m$}
\EndFor
\State $\bm{S}_{\text{concat}} \gets [\bm{S}_{\text{llm}}, \bm{S}_{\text{kg}}, \bm{S}_{\text{loc}}]$ \Comment{Concatenate scaled embeddings}
\State $\bm{X} \gets N(\bm{S}_{\text{concat}})$ \Comment{Normalize fused representation}

\Statex \textit{ --- Step 2: Evaluate via Cross-Validation ---}
\State $f_{\text{cv}} \gets 0$
\For{$i \in \{1,\ldots,k\}$} \Comment{$k$-fold Cross-Validation loop}
    \State $(\bm{X}_{\text{train}, i}, \bm{y}_{\text{train}, i}), (\bm{X}_{\text{val}, i}, \bm{y}_{\text{val}, i}) \gets$ Get fold $i$ split from $(\bm{X}, \bm{y})$
    \State $\mathcal{C}.\texttt{fit}(\bm{X}_{\text{train}, i}, \bm{y}_{\text{train}, i})$ \Comment{Fit classifier on training part of fold $i$}
    \State $\bm{y}_{\text{val}, i}^{\text{pred}} \gets \mathcal{C}.\texttt{predict}(\bm{X}_{\text{val}, i})$ \Comment{Predict on validation part of fold $i$}
    \State $f_i \gets \text{Macro-F1}(\bm{y}_{\text{val}, i}, \bm{y}_{\text{val}, i}^{\text{pred}})$ \Comment{Calculate Macro F1-score for the fold}
    \State $f_{\text{cv}} \gets f_{\text{cv}} + f_i$
\EndFor
\State $f \gets f_{\text{cv}} / k$ \Comment{Mean CV F1-macro score}

\State $\mathcal{C}.\texttt{fit}(\bm{X}, \bm{y})$ \Comment{Re-fit on the full dataset $\bm{X}, \bm{y}$}

\Statex
\Return $\mathcal{C},\, f$ \Comment{Return the final fitted classifier and the CV score}
\end{algorithmic}
\end{algorithm}

Algorithm~\ref{alg:FuDoBa} details the complete procedure for generating a fused representation $\bm{X}(\bm{\theta})$ given a specific hyper-parameter vector $\bm{\theta}$, and subsequently evaluating its quality by computing the 5-fold cross-validation macro F1-score of an AutoGluon~\cite{agtabular} model, denoted as $f(\bm{\theta})$. This score serves as our objective function, which we try to maximize, conditioned on the input parameters $\theta$.

\subsection{Bayesian Optimisation for Optimal Fusion Parameters}
\label{sec:bayesian_optimisation_revised}

Having defined the parametrized fusion mechanism and the objective function $f(\bm{\theta})$ based on cross-validation performance of the fitted AutoGluon model (Algorithm~\ref{alg:FuDoBa}), our goal is to find the optimal hyper-parameters $\bm{\theta}^*$ that maximize this objective: $\bm{\theta}^* = \arg\max_{\bm{\theta} \in \Theta} f(\bm{\theta})$. Since evaluating $f(\bm{\theta})$ is computationally expensive, we employ Bayesian optimisation (BO)~\cite{bo_nips} to efficiently search the hyper-parameter space $\Theta$. BO iteratively builds a probabilistic surrogate model of the objective function and uses an acquisition function to guide the selection of subsequent hyper-parameters to evaluate. We model $f(\bm{\theta})$ using a Gaussian Process (GP) prior~\cite{Rasmussen2006Gaussian}:
\begin{equation*}
\label{eq:gp_prior_bo}
f(\bm{\theta}) \sim \mathcal{GP}\bigl(\mu(\bm{\theta}), k(\bm{\theta}, \bm{\theta}')\bigr),
\end{equation*}
where $\mu(\bm{\theta})$ is the mean function and $k(\bm{\theta}, \bm{\theta}')$ is a kernel function, in our case the Mattern Kernel~\cite{matern}. Given $n$ observations of previous hyper-parameter evaluations and their corresponding cross-validation Macro-F1-scores, $\mathcal{D}_n = \{(\bm{\theta}_i, f(\bm{\theta}_i))\}_{i=1}^{n}$, the GP yields a posterior predictive distribution $P(f(\bm{\theta}) | \mathcal{D}_n) = \mathcal{N}(\mu_n(\bm{\theta}), \sigma_n^2(\bm{\theta}))$. Let $f^* = \max_{1\le i \le n} f(\bm{\theta}_i)$ be the current best observed cross-validation Macro-F1-score. We use the Expected Improvement (EI) acquisition function to select the next point:

\begin{equation*}
    \label{eq:ei_bo}
    \text{EI}(\bm{\theta}) = \underbrace{(\mu_n(\bm{\theta})-f^*)\,\Phi\!\left(Z\right)}_{\text{exploitation}} + \underbrace{\sigma_n(\bm{\theta})\,\phi\!\left(Z\right)}_{\text{exploration}}, \quad \text{where } Z = \frac{\mu_n(\bm{\theta})-f^*}{\sigma_n(\bm{\theta})+\epsilon}.
\end{equation*}
Here, $\mu_n(\theta)$ is the predicted mean of $f(\theta)$ based on the current observations $\mathcal{D}_n$, while $\sigma_n(\theta)$ is the predicted standard deviation, representing the uncertainty of that prediction. $\Phi$ and $\phi$ denote the standard normal cumulative distribution function (CDF) and probability density function (PDF), respectively, and $\epsilon$ is a small constant added for numerical stability. The Expected Improvement (EI) acquisition function balances exploitation—selecting points with high predicted performance—and exploration—selecting points with high uncertainty.

The next hyper-parameter configuration to evaluate is chosen by maximising the acquisition function:
\begin{equation*}
\label{eq:next_point_bo}
\bm{\theta}_{n+1} = \arg\max_{\bm{\theta} \in \Theta}\,\text{EI}(\bm{\theta}).
\end{equation*}
This iterative BO procedure allows for efficient exploration of the complex interplay between projection dimensions and importance weights, guiding the search towards an optimal fusion strategy.

\section{Experimental Setting}\label{sec:exp}
In this Section, we detail the experimental setting: the datasets and the experimental setup are  described in Section~\ref{sec:dataset} and Section~\ref{sec:expset}, respectively.

\subsection{Datasets}
\label{sec:dataset}
Following similar evaluation strategies as~\citet{babelfusion}, we evaluate our proposed method on six distinct datasets across two classification tasks: sentiment analysis and news genre classification. The sentiment analysis datasets include the Amazon Reviews collection (specifically, Books, DVD, and Music subforums) and a hate speech detection dataset comprising social media posts~\cite{ranasinghe-zampieri-2020-multilingual}. The news genre datasets are MLDoc~\cite{mldoc} (four genres) and the more recent XGENRE~\cite{kuzman-etal-2022-ginco} (nine genres). Four datasets involve binary sentiment classification, while two address multi-class news categorisation. All datasets are in English. For consistency and reproducibility, all experiments utilize the original train-test splits accompanying each dataset. Table~\ref{tab:data} provides detailed statistics for each dataset, including the number of train/test documents and average word counts.

\begin{table}[ht] 
\centering
\caption{Overview and statistics of the datasets used, similarly to ~\cite{babelfusion}}. 
\label{tab:data} 
\begin{tabular}{@{} llcrrr @{}} 
    \toprule
    \textbf{Dataset} & \textbf{Domain} & \textbf{Classes} & \textbf{Train Size} & \textbf{Test Size} & \textbf{Avg. Length (Words)} \\
    \midrule
    Books & Sentiment & 2 & 2000 & 2000 & 155.80 \\
    Dvd & Sentiment & 2 & 2000 & 2000 & 161.29 \\
    Music & Sentiment & 2 & 2000 & 2000 & 130.12 \\
    HateSpeech~\cite{ranasinghe-zampieri-2020-multilingual} & Sentiment & 2 & 13240 & 860 & 22.85 \\ 
    MLDoc~\cite{mldoc} & News & 4 & 11000 & 4000 & 235.15 \\ 
    XGENRE~\cite{kuzman-etal-2022-ginco} & News & 9 & 1650 & 272 & 1256.92 \\ 
    \bottomrule
\end{tabular}
\end{table}

\subsection{Experimental Setup}\label{sec:expset}
We design comprehensive experiments to evaluate knowledge-enriched LLM representations and analyse the effectiveness of our low-dimensional fusion strategies.

\textbf{End-to-end Classification} We compare high-dimensional LLM-only representations against knowledge-enriched versions (incorporating local, global, or combined modalities). An AutoML learner trains directly on these feature representations without preliminary dimensionality reduction. This experiment addresses the following Research Questions (RQs):
\begin{itemize}
    \item \textbf{RQ1}: Do LLM-based representations benefit from local and global knowledge enrichment?
    \item \textbf{RQ2}: What is the relative contribution of local versus global knowledge when beneficial?
\end{itemize}

\textbf{Low-dimensional Learning} We assess the efficacy of fusing learned low-dimensional, weighted, modality-specific representations before training. We benchmark this approach against direct training on the original high-dimensional concatenated representations. This experiment addresses the following RQs:
\begin{itemize}
    \item \textbf{RQ3}: Can fused, low-dimensional weighted embeddings match or exceed performance on high-dimensional features?
    \item \textbf{RQ4}: Are joint importance and dimensionality of modality optimisation superior to sequential concatenation-then-projection approaches?
\end{itemize}
\textbf{Fusion Strategies}
We evaluate two low-dimensional fusion strategies:
\begin{itemize}
    \item \textbf{FuDoBa} (our default approach): It optimises modality importance weights ($\alpha_m$) and projection dimensions ($l_m$) for each modality, before concatenating and training on the combined dimension.
    \item \textbf{Concat-then-Project (FuDoBa-CP)}: Follows the methodology in \citet{babelfusion}, where high-dimensional modality representations are first concatenated, then projected to lower dimensions before training. We fix the projected dimension to $32$.
\end{itemize}
    
\textbf{Training Protocol} To ensure fair comparisons across the different settings, the same AutoML classifier (AutoGluon) is trained on each type of feature representation, regardless of their dimensionality or fusion strategy. For each configuration of FuDoBa, we perform 50 independent Bayesian optimisation trials and estimate the 5-fold cross-validation F1-score. Each trial is limited to a runtime of 5 minutes on a commodity machine. This design emphasises our commitment to a \textit{compute-efficient} methodology for evaluating the effects of modality enrichment and fusion strategies in downstream training, as opposed to relying on costly re-training of LLM-based models~\cite{tennenholtz-etal-2024-embedding}. 

\section{Results} \label{sec:results}
We proceed with the presentation of our main results. Section~\ref{subs:e2e} details the downstream performance results, while Section~\ref{subs:abl_dims} examines the impact of dimensionality and modality importance and Section~\ref{subs:abl_kg} analyses the coverage and the extracted graph. Additional in-depth results provided in the Appendix.

\subsection{End-to-end classification}\label{subs:e2e}
\begin{figure}[!hb]
    \centering
    \resizebox{\textwidth}{!}{\includegraphics{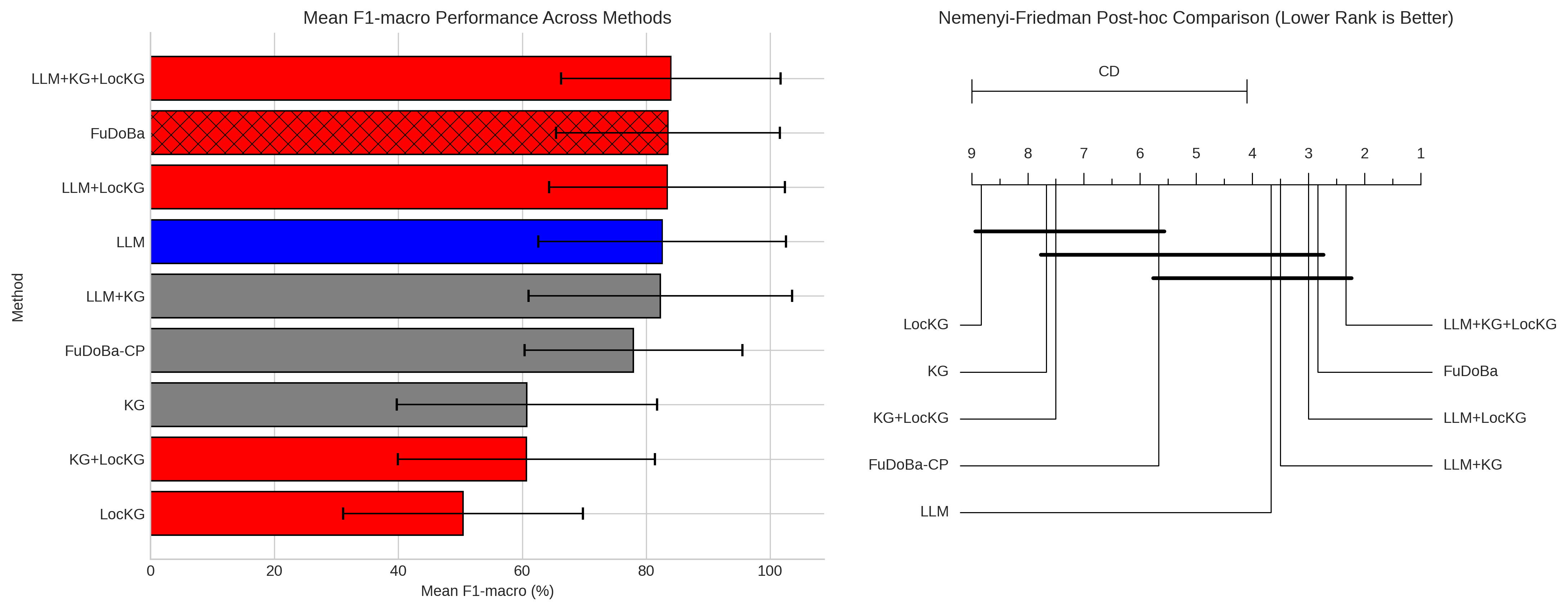}}
\caption{Average F1-score and ranking performance across datasets for AutoGluon-trained high-dimensional multi-modal representations versus the proposed FuDoBa method. The left panel shows that FuDoBa outperforms the strong LLM baseline (blue), ranking second to its high-dimensional variant. Representations in red incorporate local knowledge, a novel contribution of this work, while those in gray leverage global knowledge from WikiData as described in~\cite{babelfusion}.}    \label{fig:average_performances}
\end{figure}

Figure~\ref{fig:average_performances} and Table~\ref{tab:main_results_sentiment}  present our end-to-end classification results. We compare baseline LLM embeddings against methods incorporating Global and/or Local KGs, evaluating both simple concatenation and our proposed FuDoBa approach.

\begin{table}[ht]
    \centering
    \resizebox{\linewidth}{!}{

    \begin{tabular}{l||ccc|| ccc|c|| ccc}
\toprule
Method & \multicolumn{3}{c||}{Modality Importance Weight} & \multicolumn{3}{c|}{Projection Dimension $l$} & Training Dimension & \multicolumn{3}{c}{Macro Scores} \\

 & $\alpha_{LLM}$ & $\alpha_{KG}$ & $\alpha_{LocKG}$ & $l_{LLM}$ & $l_{KG}$ & $l_{LocKG}$ & $l_{final}$ & F1 (\%) & Prec. (\%) & Rec. (\%) \\
\midrule
\multicolumn{11}{c}{\textbf{Books}} \\
\midrule
LLM & $\checkmark$ & $\times$ & $\times$ & 1536 & $\times$ & $\times$ & 1536 & 92.80 & 92.82 & 92.80 \\ \hline
LLM+LocKG & $\checkmark$ & $\times$ & $\checkmark$ & 1536 & $\times$ & 512 & 2048 & 92.50 & 92.51 & 92.50 \\
LLM+KG & $\checkmark$ & $\checkmark$ & $\times$ & 1536 & 512 & $\times$ & 2048 & 93.25 & 93.26 & 93.25 \\
LLM+KG+LocKG & $\checkmark$ & $\checkmark$ & $\checkmark$ & 1536 & 512 & 512 & 2560 & 92.85 & 92.90 & 92.85 \\ \hline
FuDoBa & 1.0 & 0.8 & 0.1 & 64 & 32 & 16 & 112 & \textbf{93.40} & \textbf{93.41} & \textbf{93.40} \\
\midrule
\multicolumn{11}{c}{\textbf{Dvd}} \\
\midrule
LLM & $\checkmark$ & $\times$ & $\times$ & 1536 & $\times$ & $\times$ & 1536 & 92.70 & 92.71 & 92.70 \\  \hline
LLM+LocKG & $\checkmark$ & $\times$ & $\checkmark$ & 1536 & $\times$ & 512 & 2048 & 93.35 & 93.35 & 93.35 \\
LLM+KG & $\checkmark$ & $\checkmark$ & $\times$ & 1536 & 512 & $\times$ & 2048 & \textbf{93.95} & \textbf{93.95} & \textbf{93.95} \\
LLM+KG+LocKG & $\checkmark$ & $\checkmark$ & $\checkmark$ & 1536 & 512 & 512 & 2560 & 93.25 & 93.26 & 93.25 \\ \hline
FuDoBa & 1.0 & 0.1 & $\times$ & 32 & 32 & $\times$ & 64 & {93.90}&{93.90} & {93.90} \\
\midrule
\multicolumn{11}{c}{\textbf{Hatespeech}} \\
\midrule
LLM & $\checkmark$ & $\times$ & $\times$ & 1536 & $\times$ & $\times$ & 1536 & 75.68 & 77.61 & 74.41 \\ \hline
LLM+LocKG & $\checkmark$ & $\times$ & $\checkmark$ & 1536 & $\times$ & 512 & 2048 & \textbf{78.25} & \textbf{77.83} & \textbf{78.72} \\
LLM+KG & $\checkmark$ & $\checkmark$ & $\times$ & 1536 & 512 & $\times$ & 2048 & 76.30 & 78.27 & 74.99 \\
LLM+KG+LocKG & $\checkmark$ & $\checkmark$ & $\checkmark$ & 1536 & 512 & 512 & 2560 & 77.43 & 78.79 & 76.41 \\ \hline
FuDoBa & 0.8 & 0.2 & 0.9 & 64 & 64 & 16 & 144 & 74.61 & 77.27 & 73.06 \\
\midrule
\multicolumn{11}{c}{\textbf{Mldoc}} \\
\midrule
LLM & $\checkmark$ & $\times$ & $\times$ & 1536 & $\times$ & $\times$ & 1536 & 96.74 & 96.77 & 96.74 \\ \hline
LLM+LocKG & $\checkmark$ & $\times$ & $\checkmark$ & 1536 & $\times$ & 512 & 2048 & 96.79 & 96.80 & 96.79 \\
LLM+KG & $\checkmark$ & $\checkmark$ & $\times$ & 1536 & 512 & $\times$ & 2048 & 96.49 & 96.52 & 96.48 \\ 
LLM+KG+LocKG & $\checkmark$ & $\checkmark$ & $\checkmark$ & 1536 & 512 & 512 & 2560 & \textbf{97.04} & \textbf{97.05} & \textbf{97.03} \\ \hline
FuDoBa & 1.0 & 0.4 & 0.70 & 64 & 16 & 16 & 96 & 96.19 & 96.21 & 96.19 \\
\midrule
\multicolumn{11}{c}{\textbf{Music}} \\
\midrule
LLM & $\checkmark$ & $\times$ & $\times$ & 1536 & $\times$ & $\times$ & 1536 & \textbf{92.90} & \textbf{92.90} & \textbf{92.90} \\ \hline
LLM+LocKG & $\checkmark$ & $\times$ & $\checkmark$ & 1536 & $\times$ & 512 & 2048 & 92.55 & 92.55 & 92.55 \\
LLM+KG & $\checkmark$ & $\checkmark$ & $\times$ & 1536 & 512 & $\times$ & 2048 & 92.25 & 92.26 & 92.25 \\ 
LLM+KG+LocKG & $\checkmark$ & $\checkmark$ & $\checkmark$ & 1536 & 512 & 512 & 2560 & 92.60 & 92.60 & 92.60 \\ \hline
FuDoBa & 0.1 & 0.3 & 0.5 & 64 & 32 & 32 & 128 & 92.70 & 92.70 & 92.70 \\
\midrule
\multicolumn{11}{c}{\textbf{Xgenre}} \\
\midrule
LLM & $\checkmark$ & $\times$ & $\times$ & 1536 & $\times$ & $\times$ & 1536 & 44.66 & 51.43 & 57.90 \\\hline
LLM+KG+LocKG & $\checkmark$ & $\checkmark$ & $\checkmark$ & 1536 & 512 & 512 & 2560 & 50.61 & 51.78 & 63.05 \\
LLM+LocKG & $\checkmark$ & $\times$ & $\checkmark$ & 1536 & $\times$ & 512 & 2048 & 46.82 & 47.28 & 60.53 \\
LLM+KG & $\checkmark$ & $\checkmark$ & $\times$ & 1536 & 512 & $\times$ & 2048 & 41.40 & 43.73 & 54.40 \\ \hline
FuDoBa & 0.8 & 0.8 & 0.8 & 64 & 16 & 32 & 112 & \textbf{50.25} & \textbf{57.35} & \textbf{63.36} \\
\bottomrule

\end{tabular}}
    \caption{Main classification results comparing the LLM baseline, simple KG/LocKG concatenation methods, and the proposed FuDoBa technique which optimises modality integration and reduces feature dimensionality. Here, LLM denotes embeddings obtained by \textit{gpt-4o-mini}. $\checkmark$ denotes inclusion without any change, and $\times$ represents exclusion or not-applicable.  \textbf{Bolded} entries represent best scores per metric.}
    \label{tab:main_results_sentiment}
\end{table}

We find that on average, LLM-enriched representations with both local and global knowledge, boost performance compared to LLM-only representations. Furthermore, we find that on average, the best performing representation is the one utilising knowledge from both, local and global setting. We find that the enriched LLM representation (LLM+KG+LocKG) high-dimensional, outperforms FuDoBa. To assess if the differences are significant, we perform a Friedman test with Nemenyi post-hoc correction as proposed by~\citet{demsar}. We find that our method performs on-par or better compared to the LLM-based representations, with both representations being in the same ranking space with no statistically significant difference between the two. Similarly, we find no statistifically significant difference between LLM-only and FuDoBa representations. While simple high-dimensional concatenation generated the best performing score, the resulting space is high-dimensional and often infeasible to store. 
Compared to that FuDoBa, operates in low-dimensional space, offering better footprint.

\begin{figure}[!ht]
    \centering
    \includegraphics[width=0.6\linewidth]{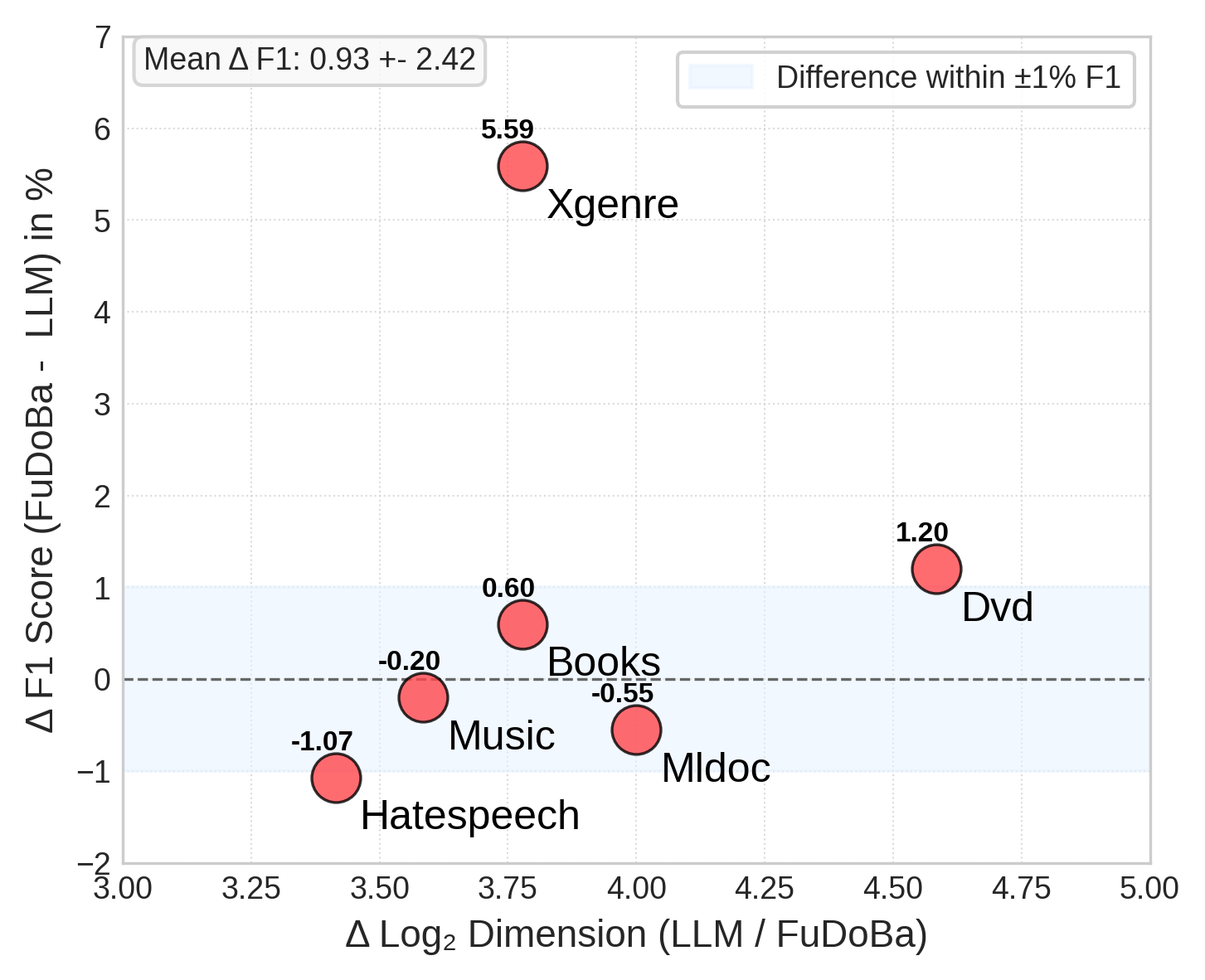}
    \caption{F1-score difference (LLM -- FuDoBa) versus the difference in log$_{2}$ of the final dimensionality. The blue shaded area indicates the region of practical equivalence, where F1-scores differ by less than one percentage point. Despite operating in lower-dimensional spaces, FuDoBa achieves results comparable or superior to those obtained with high-dimensional representations.}
    \label{fig:delta_comp}
\end{figure}

Finally, Figure~\ref{fig:delta_comp} plots the F1 performance difference (\textit{FuDoBa} - LLM) against the relative dimensionality ($\Delta \log_2$ Dimension: LLM / BF). From the visualisation we see that FuDoBa representations either score within the 1\% region of equivalence or tend to score  This visualisation confirms that \textit{FuDoBa} achieves practically equivalent (within 1\% F1) or significantly better performance (Xgenre) compared to the LLM while operating in a substantially lower-dimensional space.

\subsection{Analysis of the Impact of Modality Importance and Dimensionality} \label{subs:abl_dims}

\begin{figure}[!ht]
    \centering
    \resizebox{0.9\textwidth}{!}{\includegraphics{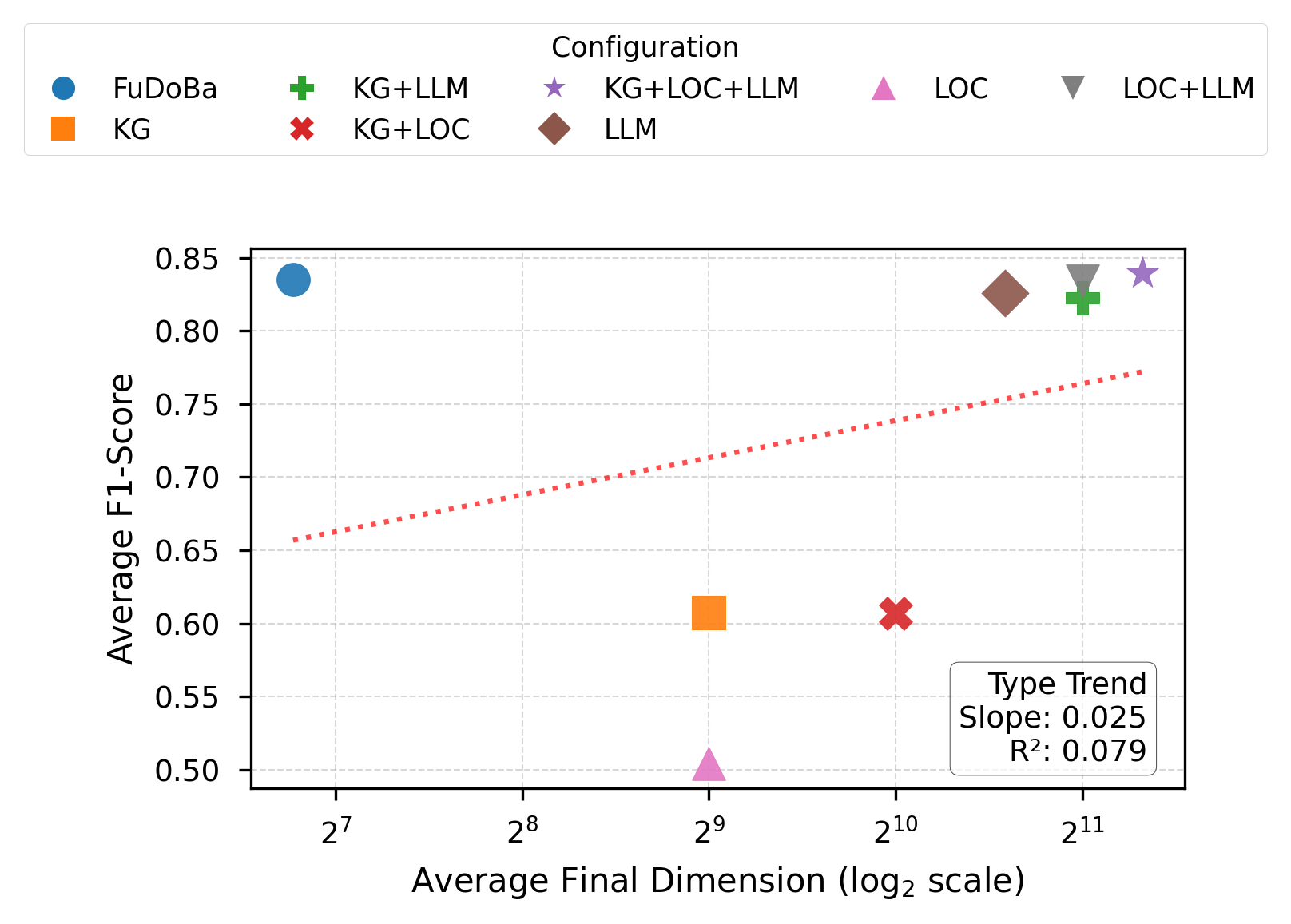}}
    \caption{The number of dimensions impact on the performance. We see a negative trend across datasets, with enriched representation methods performing on-par or better than fully fitted models, and out-performing the LLM only baselines. FuDoBa performs optimal if simultaneously considering F1-score and embedding dimension (computational complexity). }
    \label{fig:initial_plot_interactions}
\end{figure}

Next, we examine how the performance of our low-dimensional, FuDoBa representations compares to the high-dimensional concatenated representation  Figure~\ref{fig:initial_plot_interactions} illustrates the relationship between F1-score and final feature dimension ($log_2$) scale) across datasets.  While regression suggests a general trend of higher dimensions correlating with higher F1-scores, \textit{FuDoBa} consistently operates at very low dimensions (approx. $2^6$ to $2^7$, see Table~\ref{tab:main_results_sentiment}). Despite this low dimensionality, \textit{FuDoBa} frequently achieves a highly competitive performance, often exceeding the general dimensionality-performance trend and sometimes matching or surpassing higher-dimensional approaches ($> 2^{10}$) (as shown in Figure ~\ref{fig:average_performances})

Given our interpretable modality weighting schema defined by parameters $\Theta$, we next analyse the selected modality importances $\alpha$ and representational dimensions $l$ across the datasets.

\begin{figure}[!ht]
    \centering
    \resizebox{0.95\textwidth}{!}{\includegraphics{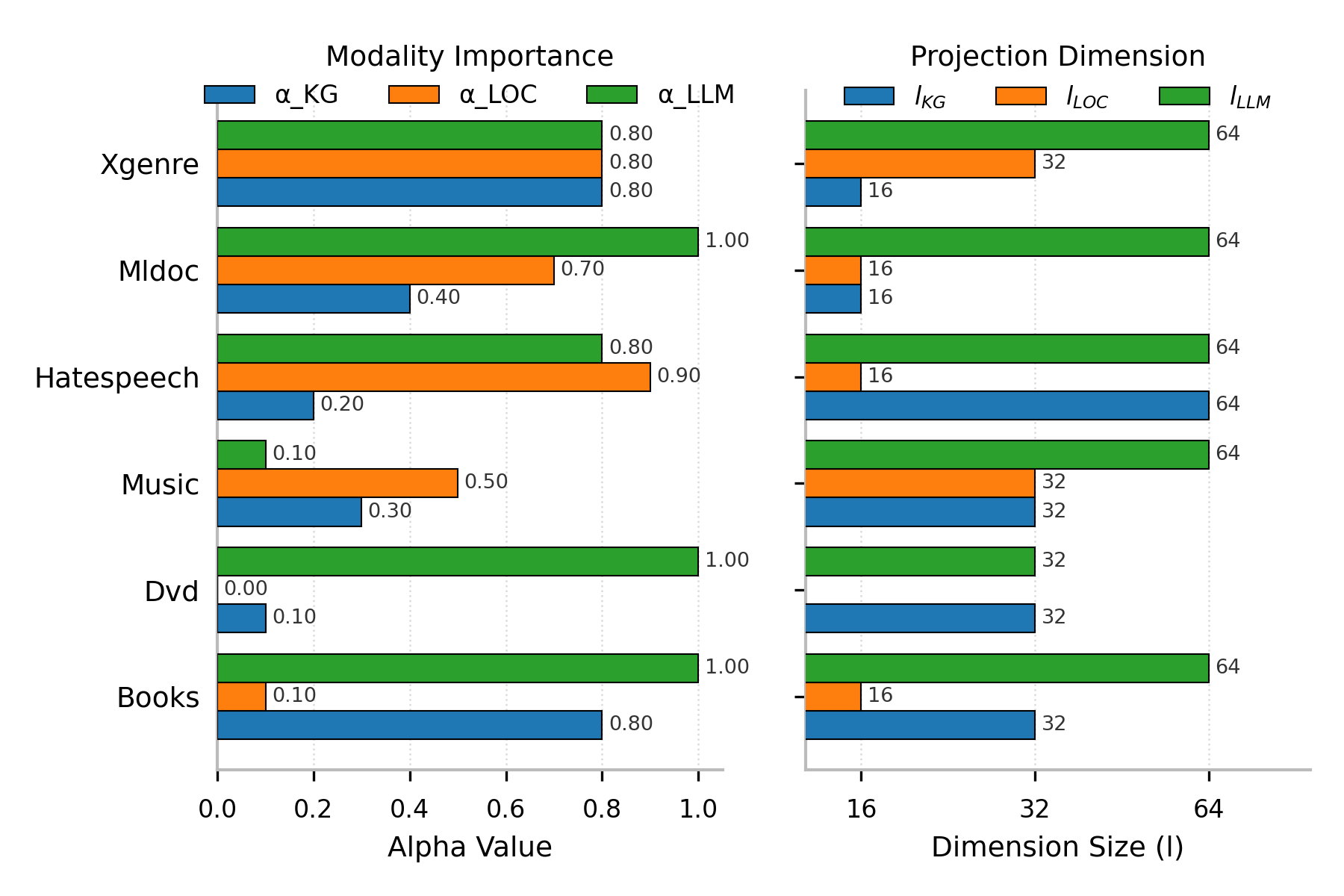}}
    \caption{Modality importance weight $\alpha$ and projection dimension $l$. It can be observed that LLM-based representation dominates -- it is most commonly selected by the Bayesian optimization as one of the key input components.}
    \label{fig:importance}
\end{figure}

In Figure~\ref{fig:importance}, the left subplot shows that modality importance, $\alpha$, varies notably across the six benchmark datasets. On average, LLM embeddings have the highest importance ($\overline{\alpha}{\text{LLM}} = 0.78$), compared to Local KGs ($\overline{\alpha}{\text{LocKG}} = 0.50$) and Global KGs ($\overline{\alpha}_{\text{KG}} = 0.43$), as expected. This may partly result from LLMs being trained on vast datasets that may overlap with the evaluation datasets. However, paired statistical tests ($N=6$) reveal no significant differences in mean importance between the modalities ($p > 0.05$ for all pairs), suggesting that modality utility is highly context-dependent and influenced by task requirements and limited sample sizes. This underscores the need for adaptive, modality-specific integration strategies. In the appendix \ref{apn:search}, we present the results of the FuDoBa search, for each dataset.

Regarding representational dimensionality ($l$, right subplot of Figure~\ref{fig:importance}), our analysis indicates distinct modality profiles. LLM embeddings consistently require higher dimensionality ($\overline{l}{\text{LLM}} = 57.33$), once again, expectedly since they stem from $1536$ dimensional dense space, significantly exceeding both Global KG embeddings ($\overline{l}{\text{KG}} = 32.00$, $p < 0.05$) and Local KG embeddings ($\overline{l}{\text{LocKG}} = 21.33$, $p < 0.01$). This likely reflects the need to capture critical variance from the dense, high-dimensional original LLM representations (1536 dimensions). Furthermore, Global KG embeddings require notably lower dimensionality for news classification tasks ($l{\text{KG}} = 16$) compared to sentiment analysis tasks ($\overline{l}_{\text{KG}} = 40.0$), while Local KG dimensionality varies but does not significantly differ from Global KG on average ($p > 0.15$). The total dimensionality of the fused feature vectors generated by \textit{FuDoBa} also varies, ranging from 80 (Dvd) to 144 (Hatespeech). News classification tasks yield slightly lower average total dimensionality ($\overline{l}_{\text{Total}} = 104$) than sentiment analysis tasks ($\overline{l}_{\text{Total}} = 116$), mainly due to lower $l_{\text{KG}}$. Despite these variations impacting computational cost, the overall dimensionality of the fused representations remains manageable compared to original LLM embedding dimensions, especially when compute-efficient training and storage is a requirement.

\subsection{Analysis of the Mapped Knowledge} \label{subs:abl_kg}
A comparative analysis of the global KG approach, inspired by the previous work~\cite{babelfusion}, and our local knowledge graph extraction method (LocKG), summarised in Table~\ref{tab:combined_stats_entities}, reveals key differences in coverage. The KG approach consistently identifies more linked entries per document, particularly in longer datasets such as XGENRE and MLDoc. In contrast, LocKG reduces the proportion of documents with no extracted information—for example, in the Hate Speech dataset, this percentage drops from 22\% (KG) to 17\% (LocKG). Additionally, LocKG yields an average of 4.99 entities per document in Hate Speech, compared to just 2.83 from the KG method.

These results highlight LocKG’s robustness in extracting meaningful information from short, unstructured texts and support the premise that local knowledge graphs offer unique advantages in handling diverse, challenging data sources. From an embedding modeling perspective, we attribute part of this improvement to the aggregation mechanism used—specifically, the averaging approach shown in Figure~\ref{fig:main-schema}. While aggregating across a larger number of entities may introduce noise, the lower noise levels in LocKG suggest that the resulting representations are more specific. That said, global KGs still capture richer entity interactions and benefit from larger, more curated training corpora. Therefore, standalone global KG embeddings are expected to remain better performing.

\begin{table}[!htbp]
 \centering
 \caption{A comparison of the extracted entities per article between the Global and Local knowledge graphs reveals that the locally constructed KG provides better coverage—fewer articles lack mapped entities. Expectedly, the local KG maps a lower number of entities per document compared to the global graph.}
 \label{tab:combined_stats_entities} 
 \begin{tabular}{@{} l rr rr rr rr @{}}
  \toprule
  \textbf{Dataset} & \multicolumn{2}{c}{\textbf{Docs w/o (\%)}} & \multicolumn{2}{c}{\textbf{Mean ± Std}} & \multicolumn{2}{c}{\textbf{Max}} & \multicolumn{2}{c}{\textbf{Median}} \\
  \cmidrule(lr){2-3} \cmidrule(lr){4-5} \cmidrule(lr){6-7} \cmidrule(lr){8-9}
                   & \textbf{KG} & \textbf{LocKG} & \textbf{KG} & \textbf{LocKG} & \textbf{KG} & \textbf{LocKG} & \textbf{KG} & \textbf{LocKG} \\
  \midrule
  Music       &  2\% &  0\% & 18.03 ± 19.27 & 15.59 ± 7.89  & 189 &  57  & 12 & 14 \\
  DVD         &  3\% &  0\% & 21.34 ± 25.08 & 17.49 ± 9.13  & 272 & 107  & 13 & 16 \\
  Books       &  3\% &  0\% & 19.00 ± 22.41 & 16.79 ± 8.40  & 203 &  78  & 12 & 15 \\
  Hate speech & 22\% & 17\% &  2.83 ±  2.63 & 4.99 ± 3.59   &  22 &  35  &  2 &  5 \\
  XGENRE      &  0\% &  0\% & 61.90 ± 48.08 & 28.34 ± 11.65 & 281 & 116  & 49 & 27 \\
  MLDoc       &  0\% &  0\% & 48.04 ± 37.01 & 24.68 ± 11.15 & 440 & 126  & 37 & 23 \\
  \bottomrule
 \end{tabular}
\end{table}

\section{Discussion}\label{sec:discussion}

While LLM-based representations alone are powerful, our results (Section~\ref{sec:results}) demonstrate that enriching them with global and local knowledge improves average downstream performance (\textbf{A1}). Furthermore, we find that low-dimensional projection of multimodal inputs can yield competitive or superior scores while significantly reducing the memory footprint (\textbf{A3}). This suggests that for use cases prioritising both performance and storage efficiency, simultaneous optimisation of per-modality dimensionality and importance can offer a practical solution for conserving resources while improving downstream results. We want to note that 50 runs and the 3 projection dimensions (16, 32, 64), might represent a particularly restrictive budget and we expect that increasing the space of dimensions and the run count would generate more promising results.
    We also want to emphasise that our proposed fusing methodology is general and can work for any set of modalities $\mathcal{M}$. For example for image and text fusion, frozen, modality-specific encoders can be used to fuse the modalities. 

From a resource perspective, this approach presents notable advantages: the optimisation approach operates on low-dimensional representations and can be performed on commodity hardware, alleviating the need for costly LLM-based fine-tuning. We also hypothesise that per-modality projection induces a form of dataset-specific embedding alignment, which may enhance AutoML model generalisation. This aspect is particularly relevant in resource-constrained academic environments, where efficient optimisation procedures can yield performance comparable to, or exceeding, that of LLM-only representations.

Our findings further indicate that no single configuration is universally optimal, even within similar domains. Optimal projection dimensions $l$ and modality importance weights $\alpha$ vary depending on the dataset and the fixed search budget (\textbf{A2}), highlighting the need for optimisation-based approaches to identify budget-constrained representations. In terms of cost-effectiveness, our method could be adapted to use pre-trained open-source relation extractors instead of LLM-based ones, with minimal impact on downstream performance for certain tasks (see Appendix), thereby further reducing computational demands. Our per-modality projection and weighting method (FuDoBa) also outperforms the concat-then-project approach (FuDoBa-CP). We hypothesise that AutoGluon’s NNs may better adapt to the manifold of concatenated low-dimensional representations, capturing both initial heterogeneity and relative importance. In contrast, the concat-then-project variant produces an orthogonal, linear SVD space in which equally weighted modalities may be over-represented in the projected fused representation, with AutoGluon models potentially amplifying this imbalance during training (\textbf{A4}).

While modality weighting can also be approached using more sophisticated NNs (e.g. via routing) or genetic algorithms (GAs), these methods often require more data and compute resources such as GPU or memory. For example, GA-based searches—such as optimising modality weights with linear classifiers \cite{autobot}—typically involve training numerous candidates over several generations, which significantly increases memory usage. We suggest that, given sufficient computational resources, our optimisation framework could potentially be adapted to parallel search strategies like GAs, though this remains to be validated.

A question might arise whether it's better to train classifier models on LLM embeddings or instead use direct LLM prompting. The embedding-based approach offers significant benefits: generating embeddings is much faster and cheaper—often 10 to 100 times less costly (see Appendix \ref{apn:comp_prices}), where we show that using text-embedding-2-small for an average 8 000-character document is up to 575 times less expensive than prompting a GPT-4 Turbo model. This also allows using less resource-intensive local models for the final classification step, as we show in this work. Furthermore, these embeddings can be reused for many different tasks later on, like unsupervised learning (i.e. clustering) or checking for shifts in the data overtime (i.e. diachronic analysis). This method can also reduce the risk of leaking sensitive data compared to prompting methods that need both input and output examples, which could reveal business secrets. However, there are drawbacks, notably the risk of data overlap where the data used to train the classifier might have already been seen by the LLM during its pre-training. On the other hand, direct LLM prompting is better at quickly adjusting to changes in the data, may need fewer examples to get started, and might perform better for certain tasks. There's also potential for synergy, where LLMs help create approximate 'soft' labels to make training data for embedding models. Despite these points, the need for powerful hardware to run large LLMs for prompting is still a major practical challenge. This highlights the value of simpler, efficient local models, which often work well with the embedding approach.

Finally, we observe that different relation extractors produce different graphs (Tables ~\ref{tab:graph-stats-restructured} and ~\ref{tab:graph-examples-restructured}), but achieve similar average scores (see Section~\ref{tab:rel_ext_comp}).  We believe this results from differences in underlying training data and architectures, which nonetheless capture domain-specific knowledge. When trained with link-completion methods such as rotatE \cite{sun2018rotate}, it seems that the resulting representations appear to normalize across extractors. While this remains a hypothesis, future analysis could explore how outputs from different extractors might be effectively combined.

\section{Conclusions and Further Work}\label{sec:conc}
In this work, we introduce FuDoBa, a framework designed to overcome the challenges of high dimensionality and costly adaptation in Large Language Model (LLM) embeddings. FuDoBa employs Bayesian optimisation to fuse core LLM representations with structured knowledge extracted from both global and local Knowledge Graphs. By adaptively learning optimal low-dimensional projections and interpretable importance weights for each modality, the framework balances information preservation with compactness, resulting in enhanced, task-specific representations.

Our experiments show that FuDoBa achieves classification performance comparable to or better than high-dimensional baselines (including LLM-only and simple concatenation approaches) while significantly reducing feature dimensionality (often using only a fraction of the original space). This demonstrates FuDoBa as a practical, compute-efficient alternative to resource-intensive LLM fine-tuning, particularly when domain adaptation is required and computing resources are limited. Moreover, our results indicate that in high-dimensional spaces, our approach can outperform LLM-only models, highlighting the benefit of incorporating external knowledge for improved performance.

For future work, we plan to apply and analyse FuDoBa on multiple domains and novel datasets. We will also compare our embedding‑based approach against direct LLM prompting to assess downstream performance differences. Because FuDoBa scales naturally to any number of modalities, we intend to investigate the integration of knowledge graphs with image data. Finally, having evaluated only English datasets to date, we will extend our experiments to additional languages.

\section*{Acknowledgments}
We acknowledge the financial support of the Slovenian Research and Innovation Agency (ARIS) through grants GC-0001 (Artificial Intelligence for Science), GC-0002 (Large Language Models for Digital Humanities), L2-50070 (Embeddings-based Techniques for Media Monitoring Applications), J5-3102 (Hate speech in contemporary conceptualizations of nationalism, racism, gender and migration) and the core research programme P2-0103 (Knowledge Technologies). The work of B.K. was supported by the Young Researcher Grant PR-12394. The work of R.N. was funded from CREATIVE project (CRoss-modal understanding and gEnerATIon of Visual and tExtual content) funded by the MUR Progetti di Ricerca di Rilevante Interesse Nazionale programme (PRIN 2020). We thank Jaya Caporusso for her feedback on an earlier draft of this manuscript.

\section*{Availability}
The complete code, including the pre-calculated embeddings, intermediate results, and scripts to reproduce the plots, will be made available upon acceptance.

\bibliography{sn-bibliography}


\begin{appendices}

\section{Limitations}~\label{abl:limitations}

Despite these promising results, FuDoBa has several limitations. Its performance relative to simpler baselines varies across datasets; for instance, in the Hate Speech dataset, a high-dimensional concatenation method outperformed FuDoBa, indicating potential sensitivity to factors such as text noise and domain variability. Additionally, the quality of the fused knowledge—particularly from the Local Knowledge Graph—depends heavily on the accuracy of the underlying relation extraction models, meaning that extraction errors can degrade the final representation. Our AutoML setup also relies on relatively short optimization runs (approximately 5 minutes), which may limit the effectiveness of the search. Finally, although FuDoBa reduces downstream training time through dimensionality reduction, the Bayesian optimization phase introduces additional computational overhead. While its sequential nature is well-suited for low-memory environments, it may be suboptimal on high-resource systems where parallel search strategies would be more efficient.

\section{Implementation Details}
~\label{abl:impl_details}
\paragraph*{Software Details}
We implemented our code using Python 3.11, organising our project with the UV package organiser. For KG embedding, we leveraged the PyKeen library configured with 512-dimensional embeddings, 1000 training epochs, and a batch size of 8192. For experimental and Bayesian hyperparameter search, we employed the Bayesian search functionality from Weights \& Biases (wandb), limiting the search to 50 runs. Additionally, we used AutoGluon 1.2 with the \texttt{good\_quality} preset and parallel fitting. The AutoGluon settings were as follows:
\begin{itemize}
  \item \texttt{fit\_strategy = 'parallel'}
  \item \texttt{num\_bag\_folds = 5}
  \item \texttt{num\_bag\_sets = 1}
  \item \texttt{time\_limit = 300 seconds}
\end{itemize}

\paragraph*{Hardware}
We evaluated all of our experiments on a standard, commodity PC.

\begin{lstlisting}[caption={System Description}, label={lst:system}]
############### CPU ##############
Architecture:             x86_64
  CPU op-mode(s):         32-bit, 64-bit
  Address sizes:          39 bits physical, 48 bits virtual
  Byte Order:             Little Endian
CPU(s):                   12
  On-line CPU(s) list:    0-11
Vendor ID:                GenuineIntel
  Model name:             Intel(R) Core(TM) i7-8700K CPU @ 3.70GHz
    CPU family:           6
    Model:                158
    Thread(s) per core:   2
    Core(s) per socket:   6
    Socket(s):            1
    Stepping:             10
    CPU max MHz:          4700,0000
    CPU min MHz:          800,0000
    BogoMIPS:             7399.70
Caches (sum of all):      
  L1d:                    192 KiB (6 instances)
  L1i:                    192 KiB (6 instances)
  L2:                     1,5 MiB (6 instances)
  L3:                     12 MiB (1 instance)
NUMA:                     
  NUMA node(s):           1
  NUMA node0 CPU(s):      0-11

############### Memory ##############
               total        used        free      shared  buff/cache   available
Mem:            62Gi        15Gi        27Gi       1,6Gi        19Gi        45Gi
Swap:          1,9Gi       1,0Gi       940Mi
\end{lstlisting}

\paragraph*{Prompt for Relation Extraction}
In our work, we used the following prompt for knowledge extraction and graph construction:
\begin{lstlisting}[caption={Prompt text for knowledge extraction and graph construction}, label={lst:prompt}]
    You are an expert in knowledge extraction and graph construction. Your task is to extract and represent information from the provided text as a collection of knowledge graph triplets in a JSON array. Each element in the array should be an object with keys "entity1", "relation", and "entity2", using the following allowed relations only: 
    IsA, PartOf, UsedFor, CapableOf, HasProperty, AtLocation, Causes, CausesDesire, Desires, MadeOf, HasSubevent, HasFirstSubevent, HasLastSubevent, NotCapableOf, NotDesires, NotHasProperty, Antonym, DefinedAs, DerivedFrom, DistinctFrom, Entails, ReceivesAction, MotivatedByGoal, CreatedBy, SymbolOf, EtymologicallyRelatedTo, FormOf, InstanceOf.
    
    Please follow these guidelines:
    1. Standardized and Unique Entities:  
       - Extract only clear, general concepts exactly as they appear in the text.  
       - Normalize entities by using lower-case and singular forms when applicable to avoid near duplicates.
    2. Concise Entities:  
       - Ensure each entity represents a single, clear concept; avoid combining multiple concepts into one entity.
    3. Robust Mapping:  
        - Do not derive or reinterpret entities-use the exact wording from the text so that each entity can be directly traced back.
    4. Simplicity in Relationships:  
       - Use the allowed relations to denote simple and clear interactions between entities.
    
    Document:
    """{document}"""
\end{lstlisting}

\section{On the impact of the Relation Extractor models}~\label{abl:rel_ext}

We next assess the impact of the Relation Extractor (RE) model on downstream task performance (Table \ref{tab:rel_ext_comp}). Within our LLM+KG+LocKG framework, we compare the performance when using the \textit{gpt4o-mini} as an extractor versus two smaller, open models (ReBeL~\cite{rebel}, ReLiK~\cite{relik}) as the RE component.

Across the six datasets, average F1-scores were highly similar for all three RE choices. Although gpt4o-mini yielded a slightly higher mean score, the overall difference between the models was marginal, spanning approximately 1.25 percentage points. A One-Way ANOVA confirmed that these minor variations in mean F1-scores are not statistically significant ($p = 0.9931$).

Therefore, the choice among gpt4o-mini, ReBeL, and ReLiK as the Relation Extractor did not significantly influence downstream performance in this experimental setting, suggesting that smaller, open-access models like ReBeL~\cite{rebel} and ReLiK~\cite{relik} serve as viable and effective alternatives to the larger, pay-per-use gpt4o-mini for this specific component within our framework. This supports the flexibility of our methodology, allowing for comparable performance with potentially lower costs and greater accessibility.

\begin{table}[h!]
\resizebox{\textwidth}{!}{

\begin{tabular}{lrrrrrr|l}
\toprule
Dataset & Books & Dvd & Hatespeech & Mldoc & Music & Xgenre & Average Perf. (\%) \\
Relation Extractor &  &  &  &  &  &  &  \\
\midrule
gpt4o-mini & 92.95 & 93.20 & 78.40 & 97.34 & 93.15 & 50.95 & 84.33 ± 17.60 \\
Rebel & 93.10 & 93.70 & 78.45 & 95.99 & 93.15 & 44.08 & 83.08 ± 20.12\\
ReLiK & \textbf{93.35} & \textbf{93.75} & 76.94 & 96.14 & 93.00 & 50.02 & 83.87 ± 17.97 \\
\bottomrule
\end{tabular}

}

\caption{Comparison of average macro F1 test scores (\%) across datasets for different relation extractors}
\label{tab:rel_ext_comp}
\end{table}

\paragraph*{Qualitative Analysis of the Extracted Relations}

Next, we conduct a qualitative examination of the extracted relations between different methods. Table~\ref{tab:graph-stats-restructured} shows significant differences in the models' extraction capabilities, both in quantity (with gpt4o-mini producing 3–8 times more triplets than its counterparts) and in the level of detail and abstractness—with gpt4o-mini generating more concise, high-level constructs like \textit{book}, \textit{government}, or \textit{movie}. On the extracted triplets side, gpt-4o-miniconsistently employs the generic "HasProperty" relation across all datasets, indicating a flexible approach to relationship categorisation. In contrast, Rebel and Relik tend toward more specific, domain-relevant relations such as "country," "performer," or "author," suggesting more structured extraction guidelines.

\begin{table}[!hb]
    \centering

    \resizebox{\textwidth}{!}{
        \begin{tabular}{lrrrrll} 
            \toprule
            \textbf{Method} & \textbf{Unique} & \textbf{Unique} & \textbf{Total} & \textbf{Unique} & \textbf{Most Freq.} & \textbf{Most Freq.} \\
            & \textbf{Ent.} & \textbf{Rel.} & \textbf{Triples} & \textbf{Triples} & \textbf{Entity} & \textbf{Relation} \\

            \midrule
            \multicolumn{7}{c}{\textbf{hatespeech}} \\ 
            \hline
            Rebel  & 8972 & 245 & 23534 & 15507 & @USER & different from \\
            Relik  & 2239 & 232 & 7341 & 5573 & @USER & country \\
            gpt-4o-mini& 14695 & 478 & 48582 & 45413 & user & HasProperty \\
            \midrule
            \multicolumn{7}{c}{\textbf{mldoc}} \\ \hline
            Rebel  & 13903 & 188 & 31139 & 21315 & United States & country \\
            Relik  & 40482 & 319 & 178950 & 105731 & its & country \\
            gpt-4o-mini& 90676 & 4269 & 251844 & 244296 & government & HasProperty \\
            \midrule
            \multicolumn{7}{c}{\textbf{music}} \\ \hline
            Rebel  & 5433 & 115 & 7279 & 6200 & The Beatles & performer \\
            Relik  & 6454 & 151 & 16511 & 13146 & this album & performer \\
            gpt-4o-mini& 17469 & 480 & 45894 & 43643 & album & HasProperty \\
            \midrule
            \multicolumn{7}{c}{\textbf{dvd}} \\ \hline
            Rebel  & 4767 & 151 & 7470 & 6102 & DVD & cast member \\
            Relik  & 5986 & 212 & 20606 & 15702 & it & cast member \\
            gpt-4o-mini& 17994 & 614 & 50721 & 48062 & movie & HasProperty \\
            \midrule
            \multicolumn{7}{c}{\textbf{books}} \\ \hline
            Rebel  & 4916 & 152 & 7046 & 5649 & Harvard University & author \\
            Relik  & 5020 & 218 & 15376 & 11453 & this book & author \\
            gpt-4o-mini& 17208 & 843 & 47746 & 45414 & book & HasProperty \\
            \bottomrule
        \end{tabular}
   } 
   \caption{Extracted Graph Statistics per Model and Dataset} 
\label{tab:graph-stats-restructured} 
\end{table}

The qualitative differences between the models shown in Table~\ref{tab:graph-examples-restructured} are even more pronounced, with each model extracting distinctly different semantic information from the same text. For example, in the mldoc dataset, Rebel extracts the factual political affiliation "(Alain Lamassoure, member of political party, UDF)," while Relik focuses on organisational structures "(France, executive body, European Commission)," and gpt-4o-minicaptures capability assertions "(France, CapableOf, more tax cuts)." This, combined with the results from Table ~\ref{tab:rel_ext_comp}, shows that despite the differences between the models—each with its own semantic focus—each captures a specific local view. Future work might consider how these specific domain views, unique to each model, could be incorporated to yield a more robust local representation. The reason for these differences likely stems from the distinct architectural and algorithmic designs of each model, with gpt4o-mini being a generalist foundational LLM, while both Rebel and Relik are specialist PLMs.

\begin{table}[!ht]
    \centering 
   
    \begin{tabular}{ll}
        \toprule
        \textbf{Model} & \textbf{Example Triple} \\ 
        \midrule
        \multicolumn{2}{l}{\textit{hatespeech}} \\ 
        
        Rebel  & (@USER, part of, BB) \\
        Relik  & (Ford, member of, conservatives) \\
        gpt-4o-mini& (she, NotDesires, leaving) \\
        \midrule
        \multicolumn{2}{l}{\textit{xgenre}} \\ \hline
        Rebel  & (Anything Brilliant, creator, Jenn Co) \\
        Relik  & (her, religion or worldview, church) \\
        gpt-4o-mini& (jenn co, instanceOf, fashion stylist) \\
        \midrule
        \multicolumn{2}{l}{\textit{mldoc}} \\ \hline
        Rebel  & (Alain Lamassoure, member of political party, UDF) \\
        Relik  & (France, executive body, Europoean Commission) \\
        gpt-4o-mini& (france, CapableOf, more tax cuts) \\
        \midrule
        \multicolumn{2}{l}{\textit{music}} \\ \hline
        Rebel  & (1956, point in time, 1956) \\
        Relik  & (this, distribution format, LP) \\
        gpt-4o-mini& (mr. farlow, DistinctFrom, wes montgomery) \\
        \midrule
        \multicolumn{2}{l}{\textit{dvd}} \\ \hline
        Rebel  & (Amazon, product or material produced, product) \\
        Relik  & (the DVD, distribution format, DVD) \\
        gpt-4o-mini& (ordered item, IsA, product) \\
        \midrule
        \multicolumn{2}{l}{\textit{books}} \\ \hline
        Rebel  & (spy, studied by, histor) \\
        Relik  & (American, ethnic group, American Indians) \\
        gpt-4o-mini& (spy histor, UsedFor, learning) \\
        \bottomrule
    \end{tabular}
     \caption{Extracted Graph Examples per Model On Same Example Per Dataset.} 
    \label{tab:graph-examples-restructured} 
\end{table}

\section{Complementary Results}\label{apn:compl}

\paragraph*{Comparison of LLM-embeddings to LLM querying prices}\label{apn:comp_prices}

We next compare the costs associated with two approaches for LLM-based tasks: direct prompting versus generating embeddings for downstream use. This analysis contrasts the estimated cost of using various proprietary (closed-source) and hosted open-source LLMs for a representative prompting task against the baseline cost of using OpenAI's text-embedding-2-small model for embedding the same input data. The base price per 1M tokens for the embedding model is $0.02\$$.

The comparison is based on the following assumptions: the prompting task involves 2000 input and 100 output tokens, while the baseline embedding task uses 2000 input tokens with text-embedding-2-small (estimated cost $\approx$ \$0.00004). All prices are based on publicly available data from service providers as of April 15, 2025.

Our findings indicate that the prompting approach is substantially more expensive than using the text-embedding-2-small baseline for embeddings. The cost ratio ranges from approximately 2.3 times higher (for Gemini 1.5 Flash) up to 575 times higher (for GPT-4 Turbo), as detailed in the subsequent analysis

\begin{table}[htbp] 
\centering
 
\resizebox{\textwidth}{!}{
    \begin{tabular}{@{}l r r r r r c@{}} 
    \toprule
    Model & Input Price & Output Price & Est. Prompt Cost & Baseline Emb. Cost & Ratio (Prompt/Emb) & Ref \\
    & (\$/1M tk) & (\$/1M tk) & (\$) & (\$) & (approx. X times) & \\
    \midrule
    \multicolumn{7}{@{}l}{\textbf{Baseline Embedding Model}} \\
    \texttt{text-embedding-2-small} & 0.02 & N/A & N/A & $\approx$ 0.00004 & 1x & \footnotemark[1] \\
    \midrule
    \multicolumn{7}{@{}l}{\textbf{Gemini Prompting Models}} \\ 
    \texttt{Gemini 1.5 Flash} & 0.0375 & 0.15 & $\approx$ 0.00009 & $\approx$ 0.00004 & $\approx$ 2.3x & \footnotemark[2] \\
    \texttt{Gemini 1.5 Pro} & 1.25 & 5.00 & $\approx$ 0.00300 & $\approx$ 0.00004 & $\approx$ 75x & \footnotemark[2] \\
    \midrule
    \multicolumn{7}{@{}l}{\textbf{GPT (OpenAI) Prompting Models}} \\
    \texttt{gpt-4o-mini} & 0.15 & 0.60 & $\approx$ 0.00036 & $\approx$ 0.00004 & $\approx$ 9x & \footnotemark[1] \\
    \texttt{gpt-3.5-turbo-0125} & 0.50 & 1.50 & $\approx$ 0.00115 & $\approx$ 0.00004 & $\approx$ 29x & \footnotemark[1] \\
    \texttt{gpt-4o} & 5.00 & 15.00 & $\approx$ 0.01150 & $\approx$ 0.00004 & $\approx$ 288x & \footnotemark[1] \\
    \texttt{gpt-4-turbo} & 10.00 & 30.00 & $\approx$ 0.02300 & $\approx$ 0.00004 & $\approx$ 575x & \footnotemark[1] \\
    \bottomrule
    \end{tabular}
 }
 \caption{Cost ratio of LLM prompting vs. \texttt{text-embedding-2-small} Embedding ($\approx$ 2000 Token Input around 8000 characters).}
\label{tab:cost_ratio_gemini_gpt}
\end{table}

\footnotetext[1]{Pricing source: OpenAI API Pricing - \url{https://openai.com/api/pricing/}}
\footnotetext[2]{Pricing source: Google AI Gemini API Pricing - \url{https://ai.google.dev/gemini-api/docs/pricing}}
\paragraph*{Feature importance between searches} \label{apn:search}

We do a meta-analysis of the importances of the $\theta$ parameters in our Bayesian search. For each dataset in Figures ~\ref{fig:books_imp} to ~\ref{fig:mldoc_imp}, we show the Bayesian evolution in the leftmost plot. Using the shaded area, we mark the best score found ($f*$ from Algorithm ~\ref{alg:FuDoBa}) so far, coupled with the feature importance—captured by fitting a Random Forest to the parameters and the achieved score—and correlation analysis in the rightmost plot. We see that across datasets, different parameters are important (e.g. for the Books dataset, the $\alpha_{kg}$ parameter seems the most important, possibly explaining why it resulted in the limit-value of $0$). While for the larger news genre dataset, we see that $l_{LLM}$ is most important. We also see that the best-value hit at different points for different datasets, with most datasets hitting it in the latter half, implying that a longer search might have resulted in better scores, probably improving the downstream performance.

\begin{center} 
    \includegraphics[width=0.9\textwidth]{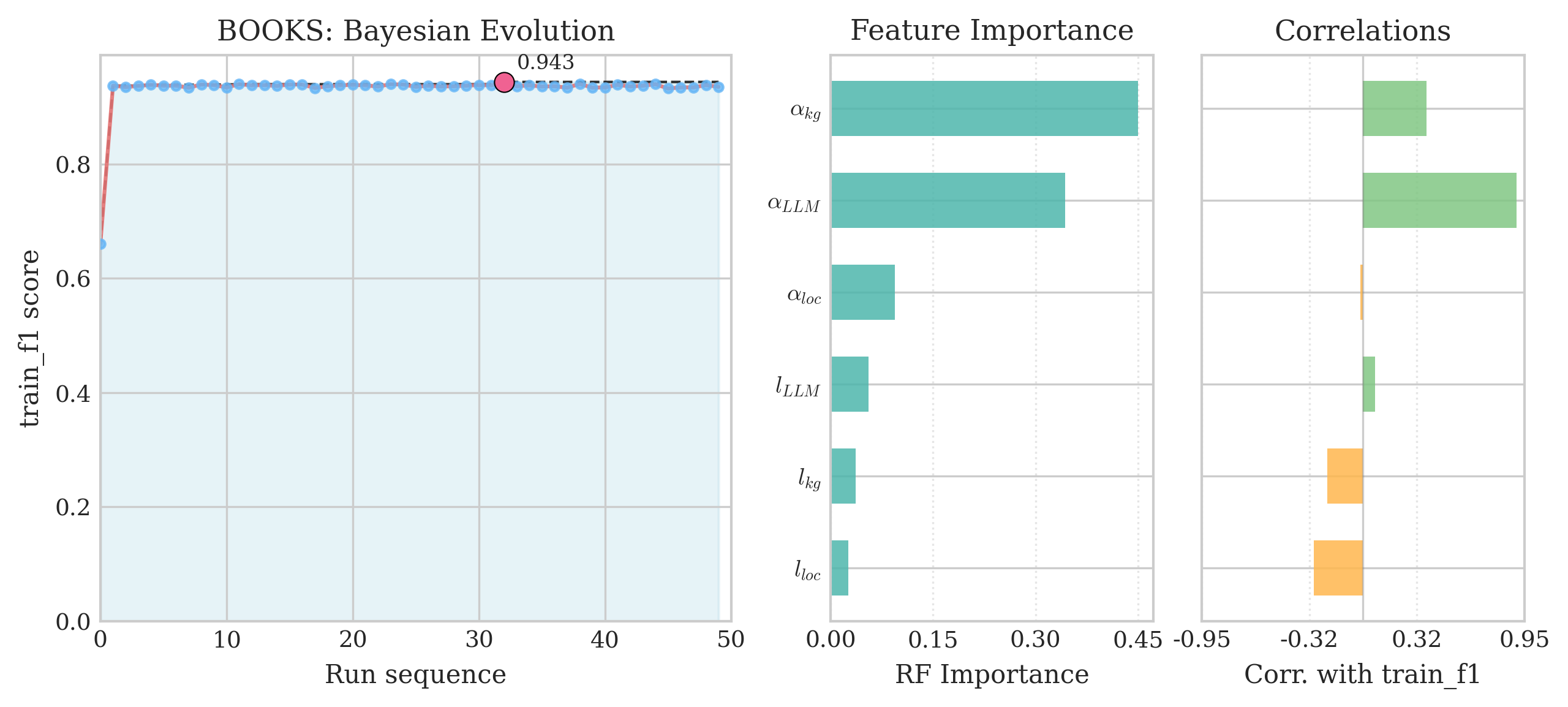}
    \captionof{figure}{Feature Importances of Bayesian Evolution for dataset Books}
    \label{fig:books_imp} 
\end{center}

\begin{center} 
    \includegraphics[width=0.9\textwidth]{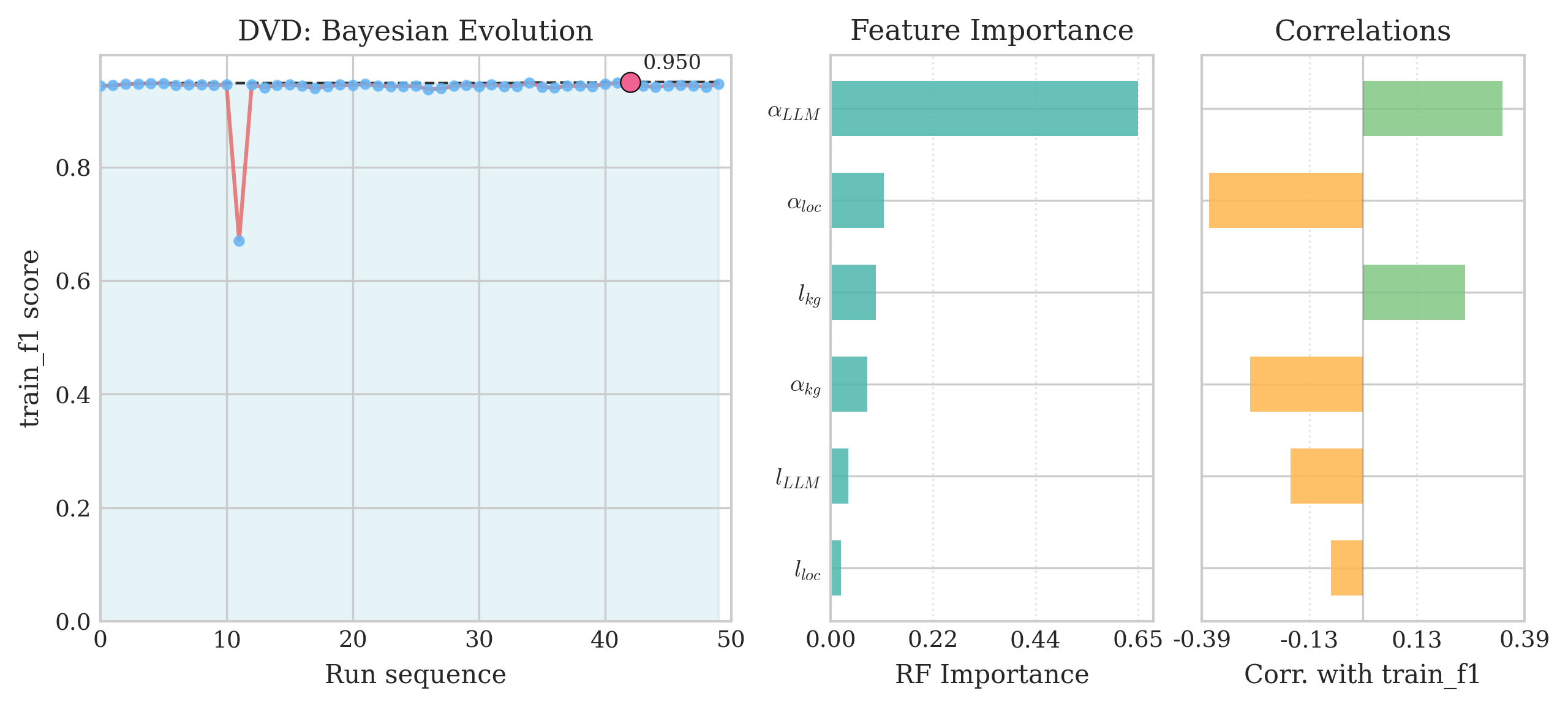}
    \captionof{figure}{Feature Importances of Bayesian Evolution for dataset Dvd}
    \label{fig:dvd_imp} 
\end{center}
\begin{center} 
    \includegraphics[width=0.9\textwidth]{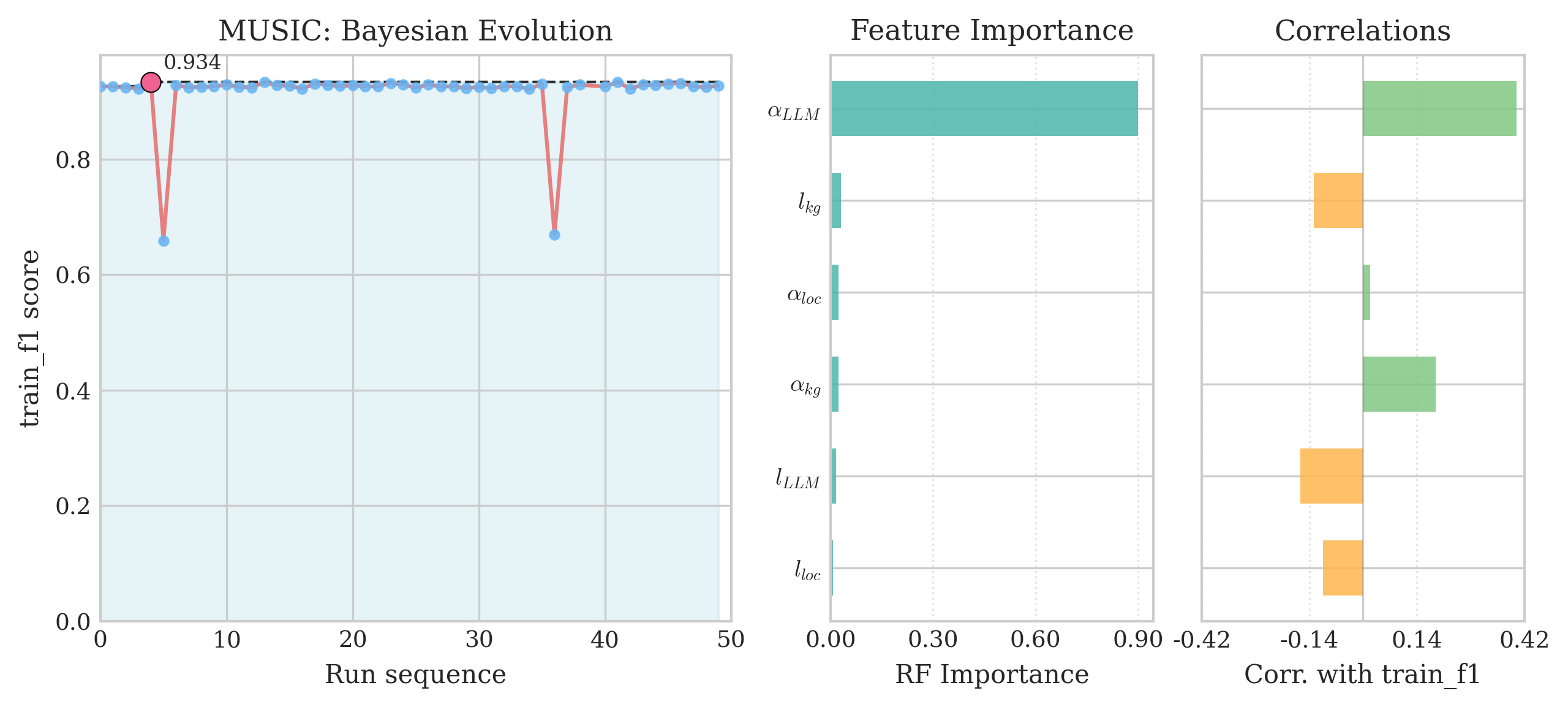}
    \captionof{figure}{Feature Importances of Bayesian Evolution for dataset Music}
    \label{fig:music_imp} 
\end{center}
\begin{center} 
    \includegraphics[width=0.9\textwidth]{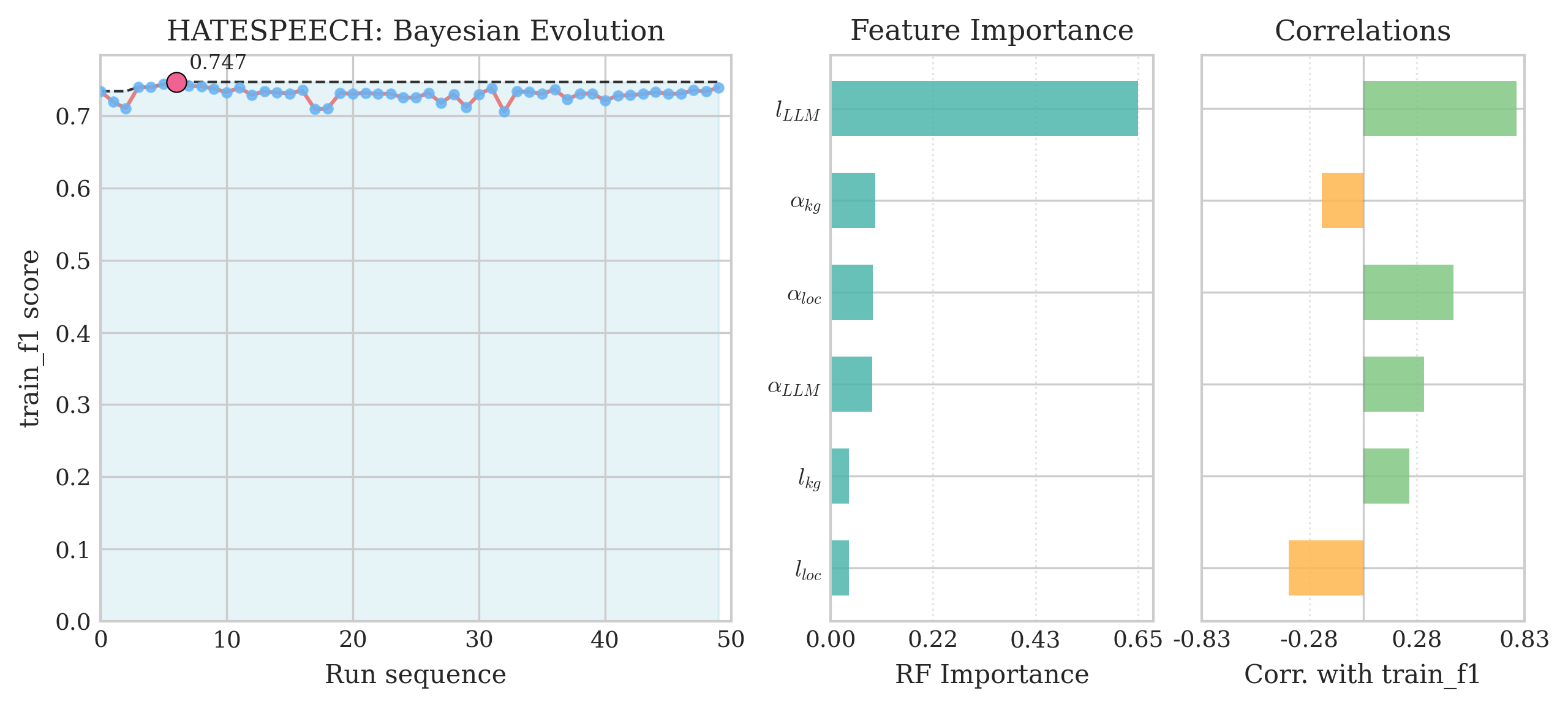}
    \captionof{figure}{Feature Importances of Bayesian Evolution for dataset Hate-speech}
    \label{fig:hs_imp}
\end{center}
\begin{center} 
    \includegraphics[width=0.9\textwidth]{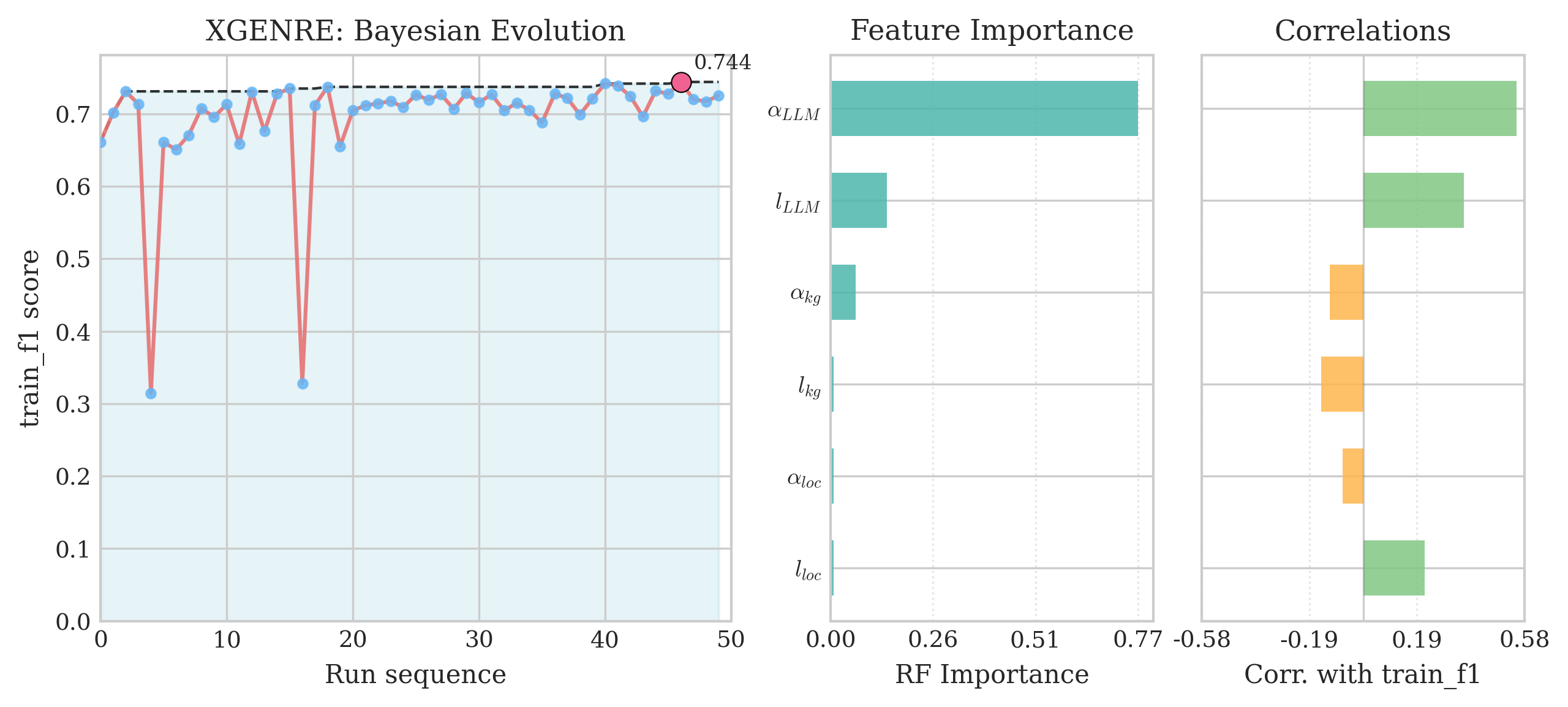}
    \captionof{figure}{Feature Importances of Bayesian Evolution for dataset Xgenre.}
    \label{fig:xgenre_imp}
\end{center}

\begin{center} 
    \includegraphics[width=0.9\textwidth]{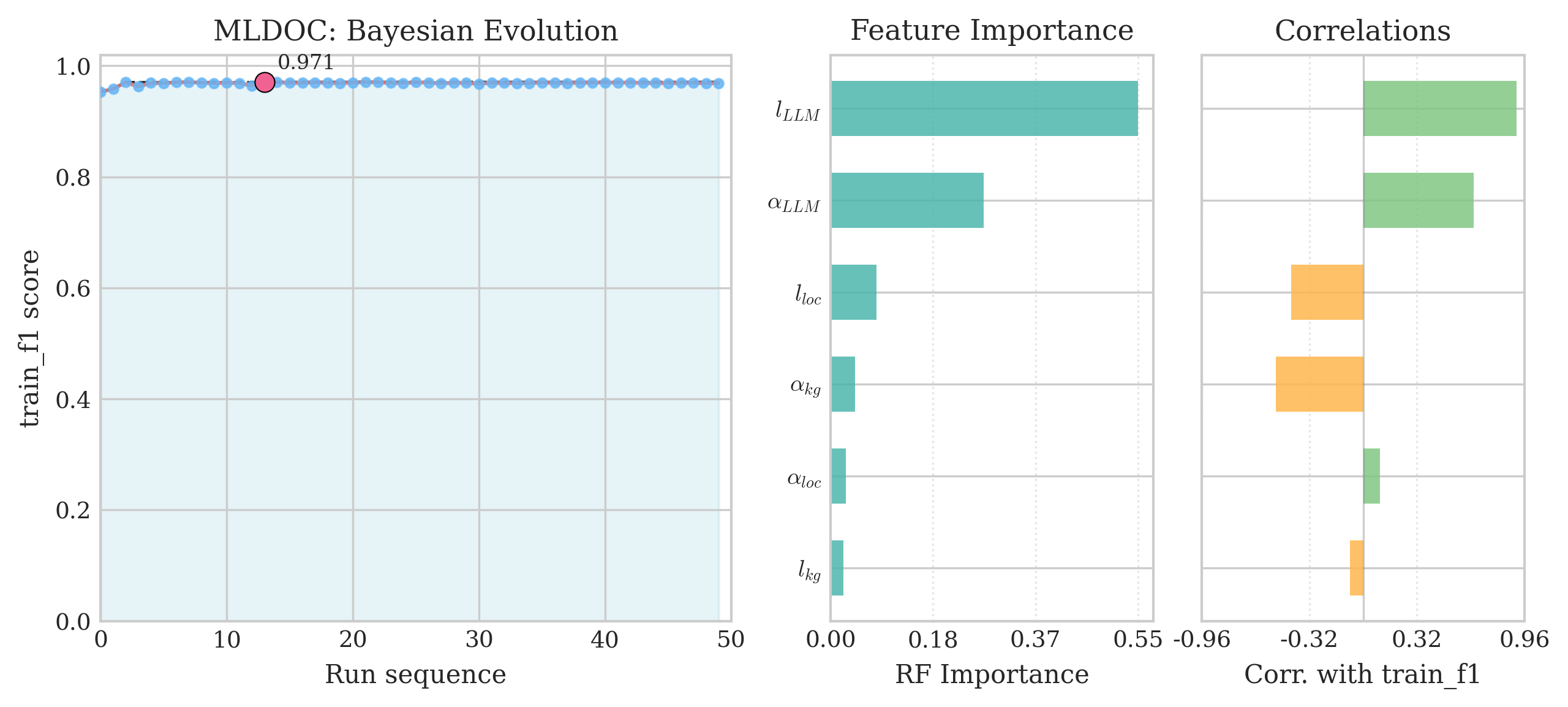}
    \captionof{figure}{Feature Importances of Bayesian Evolution for dataset MLdoc.}
    \label{fig:mldoc_imp}
\end{center}

\paragraph*{Dimension to Score Results}
Here, we present the relationship between the final projection dimension and F1-score across different datasets for each approach. Consistent with the averaged results, our approach (blue circle, FuDoBa) achieves scores comparable to—or better than—the LLM-only representation (red diamond). In general, higher-dimensional representations tend to yield better F1-scores, although the R-scores exhibit negligible trends.

\begin{figure}[ht]
    \centering
    \resizebox{\textwidth}{!}{\includegraphics{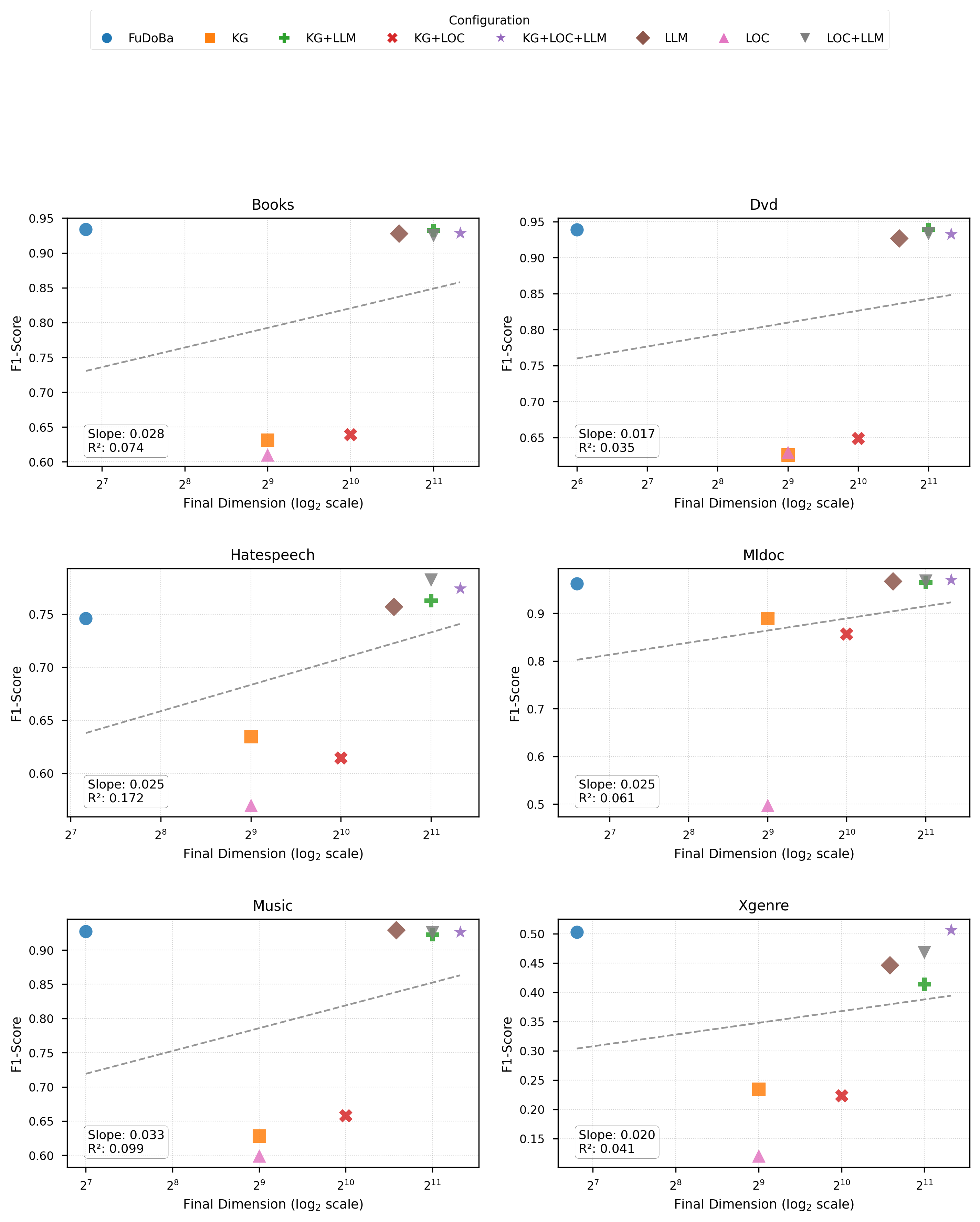}}
    \caption{Per-dataset analysis on the impact of dimensionality (x-axis, log2-scaled) and the achieved scores (y-axis, macro F1-score).}
    \label{fig:comparison_huge}
\end{figure}
\end{appendices}

\end{document}